%% file: main.tex
\documentclass{article}
\usepackage[T1]{fontenc}
\usepackage{iclr2027_conference,times}
\iclrfinalcopy
\usepackage{amsmath,amssymb,amsthm}
\usepackage{graphicx,booktabs,tabularx,array,longtable}
\usepackage{microtype,xcolor,url,placeins,float}
\usepackage{hyperref}
\definecolor{PaperLink}{HTML}{245E87}
\definecolor{SourceTone}{HTML}{596875}
\hypersetup{colorlinks=true,linkcolor=PaperLink,citecolor=PaperLink,urlcolor=PaperLink,
 pdftitle={What Must Replay Preserve? Separating Correctable Bias from Class Correspondence},
 pdfauthor={BoRen Deng, Xiangyue Ma, Chenglong Li, Xiaoting Du},bookmarksnumbered=true,pdfstartview=FitH}
\title{What Must Replay Preserve?\\Separating Correctable Bias\\from Class Correspondence}
\input{authors.tex}
\begin{document}
\maketitle
\raggedbottom
\newcommand{\vDelC}{15.2}
\newcommand{\vDelD}{23.7}
\newcommand{\vDelT}{8.4}
\newcommand{\vRmeanC}{0.33}
\newcommand{\vGenC}{0.54}
\newcommand{\vFixTwoA}{0.44}
\newcommand{\vFixTwoB}{0.08}
\newcommand{\vFixFourA}{0.03}
\newcommand{\vFixFourB}{-0.02}
\newcommand{\vFixOneA}{0.19}
\newcommand{\vFixOneB}{0.56}
\newcommand{\vFixZeroA}{0.80}
\newcommand{\vFixZeroB}{0.66}
\newcommand{\vFixTwoD}{0.16}
\newcommand{\vFixOneD}{0.05}
\newcommand{\vFixFourD}{-0.30}
\newcommand{\vFixTwoT}{-0.65}
\newcommand{\vRmeanT}{-0.31}
\newcommand{\vFixTwoMa}{-0.28}
\newcommand{\vFixTwoMb}{0.12}
\newcommand{\vHisoC}{3.4}
\newcommand{\vHpermC}{4.7}
\newcommand{\vHpermD}{4.6}
\newcommand{\vHpermT}{6.1}
\newcommand{\vNpermC}{-0.01}
\newcommand{\vNupC}{0.13}
\newcommand{\vHnatC}{5.1}
\newcommand{\vHmidC}{2.6}
\newcommand{\vHdnC}{1.1}
\newcommand{\vHbGoldA}{1.8}
\newcommand{\vHbOtherA}{3.7}
\newcommand{\vHbGoldB}{1.8}
\newcommand{\vHbOtherB}{3.9}
\newcommand{\vXFlat}{6.4}
\newcommand{\vXNone}{6.0}
\newcommand{\vXFlatNone}{-0.3}
\newcommand{\vDelCA}{16.8}
\newcommand{\vDelCC}{-4.7}
\newcommand{\vHpermCA}{4.7}
\newcommand{\vHpermCC}{12.8}
\newcommand{\vDelCLt}{1.77}
\newcommand{\vDelCLw}{-0.17}
\newcommand{\vHpermCLt}{0.13}
\newcommand{\vHpermCLw}{0.39}
\newcommand{\vDelCold}{17.8}
\newcommand{\vDelClatest}{-8.9}
\newcommand{\vHpermCold}{5.4}
\newcommand{\vHpermClatest}{-1.9}
\newcommand{\vDelClatestAbs}{8.9}
\newcommand{\vHpermClatestAbs}{1.9}
\newcommand{\vAccFullC}{37.6}
\newcommand{\vAccDelC}{22.5}
\newcommand{\vAccRmeanC}{37.3}
\newcommand{\vAccGenC}{37.1}
\newcommand{\vIntC}{5.8}
\newcommand{\vIntT}{6.4}
\newcommand{\vIntCB}{-1.4}
\newcommand{\HNMcompanion}{a companion comparison targets $e_H$ on $N$, reassigning within blocks}
\newcommand{\HNMabstract}{This correspondence cost persists under residual-energy controls and recurs with DER and on TinyImageNet.}
\newcommand{\HNMintro}{}
\newcommand{\HNMcontrib}{ and remains nonzero (1.1 pp) after its residual energy is lowered to that of $N$}
\newcommand{\HNMfigcaption}{Reassignment cost after rescaling each record's residual to the energies of Section~\ref{sec:replay-correspondence-design}; the $H$ points and the $N$ point at $e_H$ come from one 12-seed cohort, in which $N$ values are reassigned within task blocks; the $N$ point at $e_N$ is the across-$N$ shift of panel (a).}
\newcommand{\HNMparagraph}{One explanation for the tolerance of $N$ is that its centered targets are small: their terminal root-mean-square residual is 0.34 logits, against 1.88 on non-gold $H$ (Appendix~\ref{app:v6-numbers}). Lowering the $H$ residual energy reduces the correspondence cost in a separate 12-seed cohort from 5.05 pp at native energy to 2.59 pp at the geometric mean and 1.05 pp at the $N$ energy, and all three 95\% intervals exclude zero (Figure~\ref{fig:correspondence}b; Appendix~\ref{app:v75-correspondence}). Raising the $N$ residuals to the $H$ energy leaves within-block $N$ reassignment within the equivalence band ($0.13$ pp, 90\% CI $[-0.49,0.74]$), while both policies remain functional (34.6\% and 34.5\%). Magnitude therefore modulates the $H$ cost but does not by itself create it. Equal energy need not give equal displacement $\|(I-P)r\|$, so these controls bound a magnitude explanation rather than equate perturbation strength.}
\newcommand{\HBsentence}{restricting the reassignment to the example's own task block costs 1.79 and 1.76 pp, and to the other blocks 3.72 and 3.86 pp, on the first eight and the remaining twelve seeds (all 95\% intervals exclude zero; Appendix~\ref{app:v77-h-blocks}); the costs are not additive and do not localize errors.}
\newcommand{\TinyReplacement}{On TinyImageNet, both replacements recover the deletion loss but, with eight seeds, do not establish equivalence.}
\newcommand{\XDNintro}{ Refreshed targets differ: when X-DER rewrites later-class scores with a model that has learned those classes, replacing the rewritten scores by their mean costs \vXFlat{} pp in all eight seeds.}
\newcommand{\XDNsettings}{}
\newcommand{\XDNimplication}{, whereas X-DER's refreshed scores exceed a flat reference by \vXFlat{} pp (Appendix~\ref{app:v8-xder})}
\newcommand{\AbstractScope}{ on CIFAR-100}
\begin{abstract}
Class-incremental learning must recognize all classes seen so far without task labels. Logit replay methods such as DER and DER++ mitigate forgetting by matching the model's past predictions on stored examples. Deleting this matching reveals its benefit, but the resulting accuracy cost cannot show whether the stored scores themselves are needed, or whether the cost survives correction of the classifier's bias toward recent classes. We propose a diagnostic framework that treats a cached prediction as temporally heterogeneous supervision: it separates classes known when an example was stored from classes learned afterward, edits each group, and evaluates every model before and after a task-level offset that leaves within-task predictions unchanged. On CIFAR-100 with DER++, suitable fixed constants replace the unrefreshed stored scores of later-learned classes within a $\pm1$ percentage-point equivalence margin, and the offset reduces the cost of deleting their matching from 14.9 to 1.8 points. Reassigning the non-gold scores of classes known at storage, which preserves their values and each task's target probability, costs 4.3 points before and 4.0 after the offset, and a parallel cost persists in image distillation. In the tested fixed-head setting, the large cost of deleting later-class matching is thus mostly correctable by this offset, whereas the smaller cost of disrupting class correspondence persists. Code and data are available at \href{https://anonymous.4open.science/r/replay-preserve-E22B/}{\nolinkurl{anonymous.4open.science/r/replay-preserve-E22B}}.
\end{abstract}

\section{Introduction}

Visual recognition systems may encounter new categories over time while still needing to recognize those learned earlier. In class-incremental learning, one classifier must distinguish all classes seen so far without a task label at test time \citep{zhou2024cilsurvey}. Experience replay mitigates forgetting by revisiting a small memory of earlier examples \citep{cite:er}, and logit replay goes further: Dark Experience Replay (DER) stores the model's prediction scores with each example and matches them during later training, and DER++ also replays the example's label \citep{cite:derpp}. These stored scores define the supervision that earlier examples provide, so designing and evaluating replay requires knowing which of their properties matter.

Prior work has examined these scores from three directions. Replay methods revise them: X-DER refreshes the scores of classes learned after storage, and A2ER corrects inconsistent past outputs \citep{bosch2022xder,kobayashi2026improvements}. Distillation studies ask which parts of a teacher's output carry information, such as target versus non-target classes \citep{cite:dkd,huang2025deepkd}. Recency-bias studies show that class-incremental classifiers favor recent classes and correct this bias at the output, in the classifier weights, or in the training loss \citep{wu2019large,zhao2020maintaining,ma2026temporal}. None of these lines, however, settles what a deletion ablation of logit replay---the accuracy lost when its matching is deleted---actually shows. First, deletion removes the stored reference and the constraint of matching it at once, so its cost cannot show that the stored values are necessary. Second, it measures the resulting classifier under its original prediction rule, so part of the cost may be task-level bias that a post-hoc correction would remove.

\begin{figure}[t]
\centering
\includegraphics[width=0.9\linewidth]{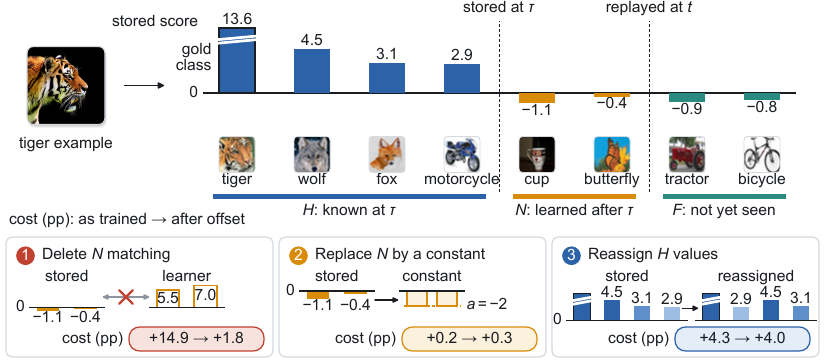}
\caption{\textbf{Different class groups require different replay information.} A stored tiger example illustrates classes known at storage ($H$), learned afterward ($N$), and unseen at replay ($F$); the photograph is illustrative. The cards show dataset-level DER++ costs of deleting $N$ matching, replacing $N$ scores with a constant, and reassigning $H$ scores, as trained and after a task-level offset (withheld-image cohort, Section~\ref{sec:v9-offset}).}
\label{fig:v6-identification}\label{fig:intro}
\end{figure}

These ambiguities matter because a cached prediction is not uniformly historical. With a preallocated output head, a record stores a score for every class, but the classes stand in different relations to the moment of storage. In the tiger example of Figure~\ref{fig:v6-identification}, the model had learned tiger, wolf, fox, and motorcycle when the example entered memory; cup and butterfly were learned only afterward, and tractor and bicycle remain unseen at replay. For the later-learned classes, does replay need the scores recorded before these classes were learned, or only the constraint of matching some reference? For classes known at storage, is it enough to keep the set of non-gold values, or must each value stay with its class? And how much of each cost survives a correction of task-level bias?

We answer these questions by replacing later-learned scores with a record mean or a constant, reassigning known-class scores within each earlier task with every value kept, and evaluating each model before and after a task-level offset that leaves within-task predictions unchanged. On ten-task CIFAR-100 with DER++, deleting matching for later-learned classes lowers final accuracy by \vDelC{} percentage points (pp), yet the constants $a=-2$ and $a=-4$ replace their stored scores within a $\pm1$ pp equivalence margin on two separate seed sets. Reassigning the non-gold scores of known classes costs a smaller \vHpermC{} pp, although the gold score and each task's values and target probability are preserved. Both costs recur under DER, on TinyImageNet, at two other memory sizes, and on ResNet-32. The offset then reverses their order: fitted on withheld training images in a separate eight-seed cohort, it reduces the deletion cost from 14.9 to 1.8 pp but leaves the reassignment cost at 4.0 pp (4.3 before). The larger as-trained cost becomes the smaller corrected cost. Under a common recipe, the same correction nearly removes ER-ACE's advantage over ER, whereas DER and DER++ retain about 5 pp; in image distillation, a parallel correspondence cost persists after correction.

Our main contributions are:
\begin{itemize}
\item \textbf{Perspective.} We treat cached logits as temporally heterogeneous supervision and distinguish three requirements that a deletion cost conflates: the matching constraint, the historical values, and class correspondence.
\item \textbf{Diagnostic framework.} We pair deletion, replacement, and reassignment with a task-level offset that separates the correctable part of each cost from the persistent part.
\item \textbf{Findings.} The deletion cost of later-class matching is neither value-specific nor mostly persistent; the cost of disrupting known-class correspondence survives correction in replay and image distillation.
\end{itemize}

\section{Related Work}

\paragraph{Replay with stored predictions.}
Experience replay trains on stored examples alongside new data \citep{cite:er,cite:icarl}, and ER-ACE changes how current and replayed examples contribute to training \citep{cite:erace}. DER and DER++ match stored logits \citep{cite:derpp}. X-DER refreshes the cached scores of classes learned after storage (its ``future past'' logits, our $N$), because those scores predate the learning of their classes \citep{bosch2022xder}. A2ER corrects inconsistent past outputs and adapts DER's objective weights \citep{kobayashi2026improvements}, and IDER enforces idempotence with respect to the previous task's model \citep{liu2026ider}. \citet{mi2025adversarial} partition logits into past, current, and future classes relative to the current stage, not to storage, to calibrate adversarial training. These methods change replay targets or training, but do not isolate whether the benefit of matching an unrefreshed cached score requires its original value.

\paragraph{Information in distillation targets.}
Knowledge distillation uses soft teacher outputs as supervision \citep{hinton2015distilling}. DKD separates target- and non-target-class contributions \citep{cite:dkd}, DeepKD further decouples and denoises their gradients \citep{huang2025deepkd}, and DIST and logit standardization change how teacher and student outputs are compared \citep{huang2022dist,sun2024logit}. Refined Logit Distillation corrects misleading teacher outputs while preserving class correlations, and a ranking-based loss emphasizes inter-class information in low-probability channels \citep{sun2025refined,guan2025kendall}. Born-Again Networks report student gains even from permuted non-maximal teacher outputs, though these trail intact outputs \citep{furlanello2018born}. In class-incremental learning, strict logit matching can conflict with new-class learning and induce recency bias \citep{gao2025fairness}; here, matching unrefreshed later-class scores instead reduces cross-task errors. We instead analyze each target by when its classes were learned relative to storage, separating the matching constraint, the stored values, and class correspondence.

\paragraph{Diagnosing class-incremental errors.}
Class-incremental prediction requires both selecting the correct task and distinguishing classes within it \citep{kim2022theoretical,lin2024tpl}. Recency bias has been corrected by rescaling new-class outputs or classifier weights \citep{wu2019large,zhao2020maintaining}, with a separated softmax \citep{ahn2021ssil}, at test time \citep{chen2025adaptive}, and by reweighting the negative supervision that earlier classes receive over time \citep{ma2026temporal}. Other work separates forgetting in representations from failure of the classifier head \citep{davari2022probing,cite:headscollapse}, and standard evaluation protocols can themselves misstate performance \citep{lai2026lie}. Our offset is a one-parameter, task-level form of prior adjustment \citep{saerens2002adjusting,menon2021logit}. We use it as a readout, not a method: its exact invariance separates errors this correction can change from those it cannot.

\section{Diagnosing What Replay Must Preserve}\label{sec:v6-design}

We propose a diagnostic framework that treats cached logits as temporally heterogeneous supervision and connects target interventions to prediction-level diagnostics. We first partition each stored vector by when its classes were learned relative to storage and decompose the matching loss on each partition (Section~\ref{sec:v6-decompose}). We then compare deletion with reference replacement to distinguish the matching constraint from the historical values it targets (Section~\ref{sec:v6-interventions}), and use value-preserving reassignment within historical task blocks to test score-to-class correspondence (Section~\ref{sec:replay-correspondence-design}). Finally, we apply a post-training task-level offset that leaves within-task predictions unchanged, measuring how much of each accuracy cost the correction recovers while evaluating within-task discrimination separately (Section~\ref{sec:replay-evaluation}).

\subsection{Class History and Matching Constraints}\label{sec:v6-decompose}

We study class-incremental learning with a classifier whose output head is preallocated for all $K$ classes. When example $(x_i,y_i)$ enters replay memory at time $\tau_i$, its logits $q_i\in\mathbb R^K$ are stored with it, forming a record $((x_i,y_i),q_i)$; $z_\theta(x_i)$ denotes the current output on that example. Our principal experiments use DER++ \citep{cite:derpp}, which combines learning on current data with historical-logit matching and replay-label supervision:
\begin{equation}
 \mathcal L
 =
 \mathcal L_{\mathrm{current}}
 +
 \frac{\alpha}{K}
 \mathbb E_{i\sim\mathcal M_1}
 \left\|z_\theta(x_i)-q_i\right\|_2^2
 +
 \beta
 \mathbb E_{i\sim\mathcal M_2}
 \operatorname{CE}\!\left(z_\theta(x_i),y_i\right),
 \label{eq:v6-objective}
\end{equation}
where $\mathcal M_1$ and $\mathcal M_2$ are independent replay draws. DER, which we also test, removes the replay-label term ($\beta=0$). The preallocated head gives every stored vector a score for each class, including classes that have not yet arrived, and the full-head cross-entropy terms push these scores down. All $K$ coordinates are matched from insertion onward, so such a class changes the temporal status of its coordinate when it arrives rather than becoming a replay target for the first time.

\paragraph{Class history.}
Let $\mathcal Y_t$ denote the classes encountered by time $t$ and $\mathcal Y^*$ the complete set of classes. At replay time $t$, we partition the coordinates of record $i$ as
\begin{equation}
 H_i=\mathcal Y_{\tau_i},\qquad
 N_{i,t}=\mathcal Y_t\setminus H_i,\qquad
 F_t=\mathcal Y^*\setminus\mathcal Y_t.
 \label{eq:v6-support}
\end{equation}
The supports contain classes encountered by insertion, encountered subsequently, and still unseen, respectively. The ground-truth class, also called the gold class, belongs to $H_i$, and coordinates move from $F_t$ to $N_{i,t}$ as their classes arrive. In an old example, $N_{i,t}$ contains scores of newer classes, whose level enters the competition between old and new tasks.

\paragraph{Matching constraints.}
We next ask what logit matching constrains on a selected coordinate set. For a nonempty coordinate block $B$ of size $n$, write $q_B=\mu_q\mathbf1+r$ and $z_B=\mu_z\mathbf1+u$, where $\mu_q,\mu_z$ are block means and $\mathbf1^\top r=\mathbf1^\top u=0$. The block contribution to Equation~\ref{eq:v6-objective} is
\begin{equation}
 \begin{aligned}
 \ell_B(z_B,q_B)
 &=
 \frac{\alpha}{K}
 \left\|(\mu_z-\mu_q)\mathbf1+(u-r)\right\|_2^2\\
 &=
 \underbrace{\frac{\alpha n}{K}(\mu_z-\mu_q)^2}_{\text{level matching}}
 +
 \underbrace{\frac{\alpha}{K}\|u-r\|_2^2}_{\text{residual matching}}.
 \end{aligned}
 \label{eq:v6-decomposition}
\end{equation}
The cross term vanishes, and the denominator stays the full head width $K$, not the block size $n$. Within residual matching, $\|u-r\|_2^2=\|u\|_2^2-2u^\top r+\|r\|_2^2$: the learner's spread and its alignment with the stored class-specific pattern are distinct contributions. The temporal partition determines where a target is changed; the loss decomposition determines which constraints it removes or retains.

\subsection{Testing Whether Stored Values Are Needed}\label{sec:v6-interventions}

The classes in $N_{i,t}$ were unseen when their scores were stored. We therefore ask whether the benefit of matching these coordinates requires the original scores or can survive a less specific reference. Full matches the cached $N$ vector, whereas Delete $N$ removes its matching penalty. Between them, Record mean keeps only each record's own mean score, and Fixed anchor uses one constant $a$ for all records (Table~\ref{tab:s3-reference-replacement}); a separately calibrated Generic anchor is defined in Appendix~\ref{app:v6-numbers}. A replacement's advantage over Delete $N$ measures the benefit retained by matching; its difference from Full measures the cost of discarded historical detail. A replacement that does not use the original $N$ values, outperforms deletion, and falls within the declared equivalence margin of Full shows that the benefit does not require those values in that setting.

\begin{table}[!htb]
\centering\small
\caption{\textbf{Target replacement retains a constraint that deletion removes.} Definitions on the support $N=N_{i,t}$. All DER++ policies keep replay labels and $H/F$ matching; $\ell_{\rm flat}$ as in Equation~\ref{eq:replay-flat-constraint}.}
\label{tab:s3-reference-replacement}
\begin{tabular}{@{}lll@{}}\toprule
Policy & Reference on $N$ & Selected-block loss\\\midrule
Full & Cached vector $q_{i,N}$ & $\ell_N(z_N,q_{i,N})$\\
Record mean & $\mu_{q_i,N}\mathbf1$ & $\ell_{\rm flat}(z_N;\mu_{q_i,N})$\\
Fixed anchor & $a\mathbf1$ & $\ell_{\rm flat}(z_N;a)$\\
Delete $N$ & None & $0$\\\bottomrule
\end{tabular}
\end{table}

A constant reference is not the absence of supervision. Replacing $q_B$ with $a\mathbf1$ in Equation~\ref{eq:v6-decomposition} yields
\begin{equation}
 \ell_{\mathrm{flat}}(z_B;a)
 =
 \underbrace{\frac{\alpha n}{K}(\mu_z-a)^2}_{\ell_{\mathrm{level}}}
 +
 \underbrace{\frac{\alpha}{K}\|u\|_2^2}_{\ell_{\mathrm{dispersion}}}.
 \label{eq:replay-flat-constraint}
\end{equation}
The first term anchors the learner's mean score; the second penalizes deviations around that mean. A flat target removes the stored residual pattern but retains a penalty on the dispersion of the learner's logits: $\nabla^2_{z_B}\ell_{\rm flat}=(2\alpha/K)I_n$ acts on all $n$ directions. When $N_{i,t}$ spans several newly learned tasks, different task-level mean scores incur dispersion loss even if scores are constant within each task.

To test whether either term alone reproduces the joint flat-target benefit, we compare four policies: Neither (that is, Delete $N$), Level only, Dispersion only, and Both, with selected-block losses $0$, $\ell_{\mathrm{level}}$, $\ell_{\mathrm{dispersion}}$, and $\ell_{\mathrm{level}} + \ell_{\mathrm{dispersion}}$, respectively. Because every retained term preserves its original coefficient and normalization, the single-term policies differ from Both only by the removal of the other component, not by its reweighting. For accuracy $J$, we measure their interaction as $I_J=J_{\mathrm{Both}}-J_{\mathrm{Level}}-J_{\mathrm{Dispersion}}+J_{\mathrm{Neither}}$. A positive value means that the joint gain exceeds the sum of the individual gains relative to Neither; it is measured across trained policies and is not implied by the additive form of the loss.

\subsection{Testing Class Correspondence}\label{sec:replay-correspondence-design}

Flat replacement changes the stored values and erases their original class-specific pattern, so it cannot isolate the effect of score-to-class correspondence. We test that property on $H_i$, whose classes had already been encountered when the record was stored, by retaining non-gold values and reassigning them within historical task blocks. In the residual term of Equation~\ref{eq:v6-decomposition}, a permutation $P$ preserves $\|Pr\|_2=\|r\|_2$ but can change the alignment $u^\top Pr$, and hence which class-specific direction is matched. Perm $H$ orders the non-gold classes within each historical task block by head index and cyclically moves each stored value to the next class; the gold coordinate stays fixed. Each block's value multiset, mean, and centered norm are therefore unchanged, as are the gold probability and each task's total probability in the cached target.

Three controls bound the interpretation. $H$ block mean replaces non-gold values by their block mean, removing correspondence by changing values rather than misassigning them. A cyclic shift across all of $N_{i,t}$ applies the same kind of reassignment to later-learned classes; because it need not preserve each task's probability mass, comparing it with Perm $H$ measures sensitivity to the two interventions, not the isolated effect of when a class was learned. Finally, to test whether magnitude alone explains a difference in cost, we rescale $H$ residuals to $e_H$, $\sqrt{e_He_N}$, or $e_N$ and $N$ residuals to $e_H$, where $e_H$ and $e_N$ are a record's mean squared centered non-gold $H$ and $N$ scores, and compare original with reassigned targets (Appendix~\ref{app:v7-energy}). All interventions leave matching on untreated coordinates and the full-width normalization unchanged. Paired policies share class order, initialization, and data order (varying across seeds), but their learners and memories may diverge.

\subsection{Separating Correctable from Persistent Costs}\label{sec:replay-evaluation}

Class-incremental classifiers tend to favor the most recent task \citep{wu2019large}, so part of any accuracy cost may be task-level bias rather than lost discrimination. Our primary endpoint $J$ is final accuracy over all seen classes, without a supplied task identity; we evaluate the same classifier before and after a correction that can change only cross-task errors.

\paragraph{Task-level offset.}
For each trained model we add one scalar $b$ to the logits of the final task's classes $T_{\rm last}$, $z'_c(x)=z_c(x)+b\,\mathbb{I}[c\in T_{\rm last}]$, choosing $b$ so that the model's mean predicted probability of $T_{\rm last}$ on the fitting images equals that task's known class share. Because $b$ shifts every logit of one task equally, predictions restricted to the final task, or to all earlier tasks pooled, are exactly unchanged; we call accuracy restricted to the true one of these two groups \emph{group-restricted accuracy}. Following \citet{kim2022theoretical}, $J=C_{\rm task}-X$, where within-task accuracy $C_{\rm task}$ restricts prediction to the true task's classes and $X$ is the fraction of examples classified correctly within their true task but assigned a class from another task; the offset changes $J$ only through $X$. For a variant $V$ of the full target with as-trained cost $U=J_{\rm Full}-J_V$, the corrected cost $U^*=J'_{\rm Full}-J'_V$ is the part that persists after this correction, and $U-U^*$ is the part it recovers. Each model receives its own offset, so both parts are readouts of trained models, not effects of matching during training. The three withheld-image cohorts fit $b$ on 500 training images withheld before training, so test images are only scored; most other cohorts use five-fold cross-fitting on the test set (Appendix~\ref{app:cal}).

\paragraph{Statistics.}
We report paired differences, called costs when the full target is the reference, with 95\% confidence intervals. For equivalence tests, we instead use 90\% paired confidence intervals: two policies are considered equivalent on $J$ only if the entire interval lies within $[-1,+1]$ pp (two one-sided tests, TOST, at the 5\% level). This equivalence margin is below the seed-to-seed standard deviation of the full target (1.2--1.7 pp in the main cohorts; Appendix~\ref{app:v6-robustness}) and is approximately 7\% of the deletion cost. Tests labeled prespecified follow internal test plans written before their cohort ran and included in the supplementary material; they are not public preregistrations.

\section{Experiments}\label{sec:v76-experiments}\label{sec:v6-experiments}

We examine whether the $N$-matching constraint and its reference are necessary (Section~\ref{sec:v76-main}), whether class correspondence matters for $H$ (Section~\ref{sec:v76-analysis}), whether both costs persist after a task-level offset (Section~\ref{sec:v9-offset}), and whether the same analysis applies beyond target interventions (Section~\ref{sec:v9-scope}). The principal study uses a preallocated ResNet-18~\citep{he2016deep} classifier trained on ten-task CIFAR-100~\citep{krizhevsky2009learning} with DER++, a memory capacity of 500 examples, and 50 epochs per task. Within each paired comparison, the policies use identical optimization settings ($\alpha=0.3,\beta=0.5$ in the principal DER++ study) and differ only in the target intervention. The main comparisons use 20 paired seeds; the prespecified offset analyses train new models on separate seeds. Follow-up experiments assess the scope of the findings on TinyImageNet~\citep{le2015tiny}, with DER instead of DER++, with memory capacities of 200 and 2{,}000 examples, and with ResNet-32 (settings in Appendix~\ref{app:v6-robustness}).

\begin{figure}[t]
\centering
\includegraphics[width=\linewidth]{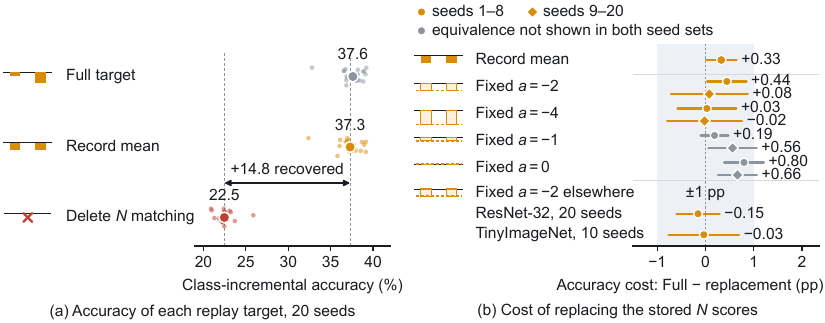}
\caption{\textbf{Deleting $N$ matching loses accuracy while simplified references retain it.} Glyphs show what the stored $N$ scores become. (a) Final accuracy over 20 paired seeds (dots) and means. (b) Full minus replacement, with 90\% paired intervals and the $\pm1$ pp equivalence band; each constant is tested separately on seeds 1--8 (circles) and 9--20 (diamonds) and is grey unless both intervals lie in the band; single-marker rows use all seeds of their cohort; ``elsewhere'' rows reuse $a=-2$ in other settings, and TinyImageNet passes only its original rule.}
\label{fig:s41-evidence}\label{fig:v7-results}\label{fig:replace}
\end{figure}

\subsection{The Benefit of \texorpdfstring{$N$}{N} Matching Does Not Require Its Stored Values}\label{sec:v76-main}

\paragraph{Deleting $N$ matching is costly, but replacing its stored values is not.}\label{sec:v7-replacement}\label{sec:v6-specificity}\label{sec:v6-anatomy}
Full, Record mean, and Delete $N$ reach 37.6\%, 37.3\%, and 22.5\% on CIFAR-100 (Figure~\ref{fig:s41-evidence}a). Deletion lowers accuracy in every seed, whereas Full exceeds Record mean by only 0.33 pp (90\% CI $[0.01,0.64]$), meeting the equivalence criterion; Record mean recovers 14.8 of the 15.2 pp. The deletion cost recurs under DER (23.7 pp), on TinyImageNet (8.4), at memories of 200 and 2{,}000 (9.1 and 11.7), and on ResNet-32 (6.3). Deletion thus shows that $N$ matching is valuable, not that its stored values are.

\paragraph{The reference can be reduced to a constant.}
Record mean removes class-specific variation but still reads one mean per record from its stored scores; a fixed anchor tests whether even this is needed. Tested separately on seeds 1--8 and 9--20, $a=-2$ and $a=-4$ cost $-0.02$ to $0.44$ pp, all four 90\% intervals lie within $\pm1$ pp, and each constant recovers more than 14 pp over deletion (Figure~\ref{fig:s41-evidence}b; Table~\ref{tab:v77-fixed-v6}). Reused unchanged, $a=-2$ meets the criterion at memories of 200 and 2{,}000, under DER (as do $a=-1$ and $-4$), and on ResNet-32 (20 seeds); on TinyImageNet, a ten-seed cohort passes its original rule but not a stricter one adopted after an interim look (Table~\ref{tab:v8-settings}). The constant is not arbitrary: $a=0$ (both seed sets) and $a=-1$ (seeds 9--20) do not establish equivalence, although both recover most of the deletion cost (Appendix~\ref{app:v77-fixed}). Because the stored $N$ scores vary little within each record (Appendix~\ref{app:v6-numbers}) and a suitable constant serves as well, $N$ matching appears to act mainly by holding later-class scores down on old examples, a cross-task effect that Section~\ref{sec:v9-offset} tests directly. This replaceability concerns unrefreshed scores: flattening X-DER's refreshed later-class scores costs \vXFlat{} pp (Appendix~\ref{app:v8-xder}).

\subsection{\texorpdfstring{$H$}{H} Matching Requires Class Correspondence Beyond Value Preservation}\label{sec:v76-analysis}

\begin{figure}[t]
\centering
\includegraphics[width=\linewidth]{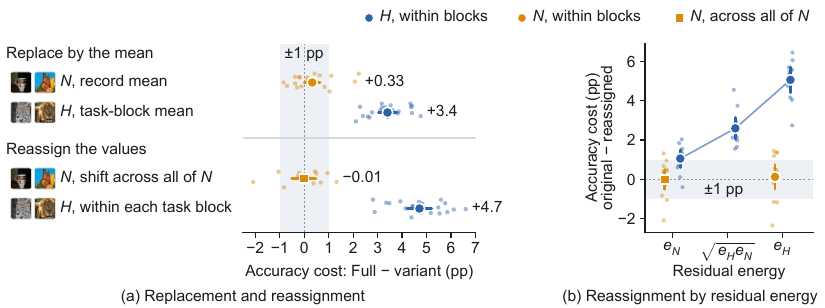}
\caption{\textbf{Changing which class holds which value is costly on $H$ and not on $N$.} Dots are seeds; whiskers are 95\% paired intervals for $H$ and 90\% for $N$, whose band is the $\pm1$ pp margin. (a) Replacing values by a mean (record mean on $N$, task-block mean on $H$; 20 seeds each) and reassigning them (a cyclic shift across $N$, 12 seeds; within each $H$ task block, 20 seeds). (b) \HNMfigcaption}
\label{fig:correspondence}\label{fig:v75-correspondence}
\end{figure}

\paragraph{Preserving values does not preserve their effect on $H$.}\label{sec:v7-correspondence}\label{sec:v6-correspondence}
The replacement that was nearly free on $N$ is costly here: replacing non-gold $H$ scores by their task-block means costs 3.4 pp on CIFAR-100 (Figure~\ref{fig:correspondence}a; separate implementation, descriptive only) and 4.1 on TinyImageNet. Keeping every value and only reassigning them within task blocks costs 4.7 pp (95\% CI $[4.2,5.2]$; positive in all 20 seeds), although the gold score, each block's values, and each task's target probability are preserved. The same reassignment costs 4.6 pp under DER, 6.1 on TinyImageNet, 2.6 and 3.7 at memories of 200 and 2{,}000, and 3.8 on ResNet-32, whereas cyclically shifting the $N$ values costs $-0.01$ pp (90\% CI $[-0.50,0.48]$), within the equivalence band. Which class holds which value therefore matters beyond the preserved values and task mass, consistent with the relational role of soft targets~\citep{hinton2015distilling}. Because a reassigned target also supplies wrong class relations, we interpret it alongside the block-mean control, which removes correspondence without misplacing it. Both the record's own task block and its other seen blocks carry the cost (Appendix~\ref{app:v77-h-blocks}).

\paragraph{Residual magnitude alone does not explain the contrast.}
In terminal memory, centered $N$ targets are smaller than non-gold $H$ targets (RMS 0.34 versus 1.88 logits), so $N$ might tolerate reassignment merely because they are small. In a separate 12-seed cohort, lowering the $H$ residuals to the $N$ energy reduces the reassignment cost from 5.05 to 1.05 pp, but both 95\% intervals exclude zero (Figure~\ref{fig:correspondence}b). Conversely, raising the $N$ residuals to the $H$ energy leaves reassignment within the equivalence band (0.13 pp, 90\% CI $[-0.49,0.74]$). Magnitude thus modulates the $H$ cost but does not explain the contrast between supports, though equal energy need not imply equal displacement $\|(I-P)r\|$.

\subsection{A Task-Level Offset Reverses the Order of the Two Costs}\label{sec:v9-offset}\label{sec:v6-behaviors}

Both costs above are measured as trained; we now ask how much of each survives the offset.

\begin{figure}[t]
\centering
\includegraphics[width=\linewidth]{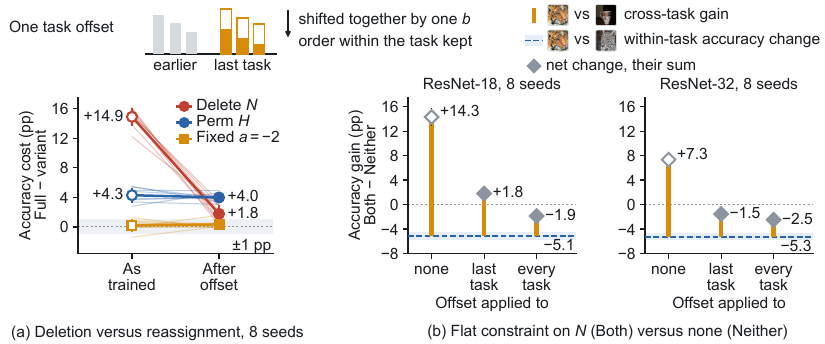}
\caption{\textbf{A task-level offset removes most of the deletion cost; the reassignment cost remains.} Hollow markers show models as trained and filled markers the same models after the offset, with 95\% paired intervals. (a) Cost relative to Full in the prespecified withheld-image cohort (lines are seeds). (b) Gain of Both over Neither with $a=-2$ (eight seeds each; offsets cross-fitted on test halves): a within-task part that no offset changes (dashed) plus a cross-task part (bars).}
\label{fig:offset}
\end{figure}

\paragraph{The offset recovers most of the deletion cost but little of the reassignment cost.}
We train Full, Delete $N$, Fixed ($a=-2$), and Perm $H$ on eight new seeds and fit each model's offset on 500 withheld training images. Deleting $N$ matching costs 14.9 pp as trained and 1.8 pp after the offset (95\% CI $[0.7,2.8]$), whereas reassigning $H$ costs 4.3 pp as trained and 4.0 pp after it ($[3.5,4.5]$; Figure~\ref{fig:offset}a). The larger as-trained cost thus becomes the smaller corrected cost: after the offset, Perm $H$ minus Delete $N$ is $-2.2$ pp ($[-3.2,-1.2]$), negative in all five cohorts of Table~\ref{tab:cal-interventions}. All three prespecified tests pass under Holm correction: the offset reduces the deletion cost, the corrected reassignment cost exceeds zero, and Fixed is equivalent to Full after the offset (TOST $p=.006$). This equivalence concerns the anchor's cost, not its advantage over Delete $N$ after correction.

\paragraph{The corrected reassignment cost persists on other backbones.}
On ResNet-32 and a narrower ResNet-18, all three tests pass again, and the corrected reassignment cost is 4.2 pp (95\% CI $[3.7,4.7]$) and 3.2 pp ($[2.2,4.2]$). The corrected deletion cost changes sign across the three configurations ($+1.8$, $-1.6$, and $-0.6$ pp; the last interval includes zero). With one offset per task, Delete $N$ is 2.4 to 3.0~pp more accurate than Full in all three, while reassignment still costs 3.2 to 4.4 pp (post hoc for ResNet-18; Appendix~\ref{app:cal}).

\paragraph{Reassignment impairs discrimination that the offset cannot change.}
In group-restricted accuracy, which no final-task offset changes, Full is 4.6 pp more accurate than Perm $H$, with 0.46 nats lower group-conditional cross-entropy; relative to Delete $N$, it is 2.0 pp more accurate with similar cross-entropy ($+0.04$ nats, $[-0.02,0.09]$). Pooling earlier tasks still allows errors between them; within the true task ($C_{\rm task}$), reassignment costs 12.4 pp, whereas Delete $N$ outperforms Full by 5.0 pp (post hoc; Table~\ref{tab:cal-backbones}). Together, these results suggest that $N$ matching primarily helps across tasks, while $H$ correspondence helps within them.

\paragraph{The joint benefit of level and dispersion is also mainly cross-task.}
With $a=-2$, Both raises accuracy over Neither by 14.3 and 7.3 pp on ResNet-18 and ResNet-32 (Figure~\ref{fig:offset}b). The gains fall to $+1.8$ and $-1.5$ pp with a final-task offset, and to $-1.9$ and $-2.5$ pp with per-task offsets (post hoc, fitted with labels), while Both costs 5.1 and 5.3 pp in within-task accuracy, which no offset changes. The interaction $I_J$ is $+4.0$ to $+7.0$ pp as trained but $-1.1$ to $+1.2$ pp after the offset (Table~\ref{tab:cal-factorial}).

\subsection{Correctable and Persistent Gains Beyond Target Interventions}\label{sec:v9-scope}

The offset may also change how replay methods compare, and correspondence may matter wherever soft targets are matched.

\begin{figure}[t]
\centering
\includegraphics[width=\linewidth]{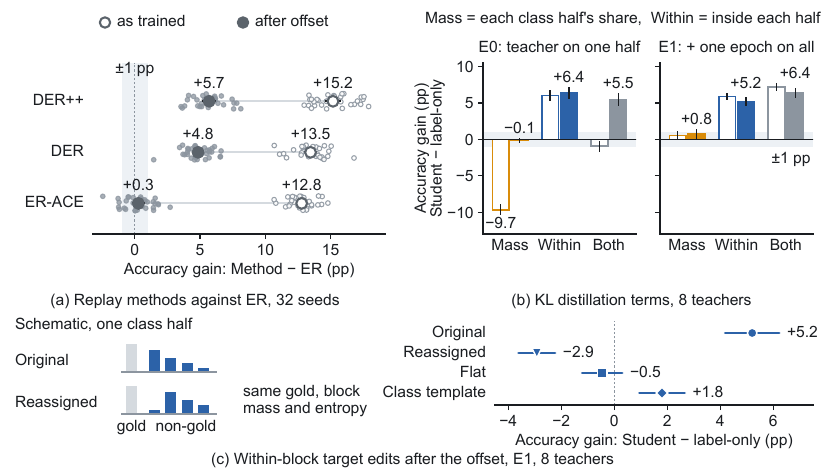}
\caption{\textbf{The offset separates correctable from persistent gains beyond target interventions.} Markers and intervals as in Figure~\ref{fig:offset}. (a) Gain over ER of three replay methods (32 paired seeds). (b) Gain of distillation students over label-only training. (c) Corrected gain over label-only training for original and edited within-block targets in a separate E1 cohort; left, schematic.}
\label{fig:scope}
\end{figure}

\paragraph{Standard replay methods.}
Under the principal offline 50-epoch recipe on 32 paired seeds, the offset reduces the advantage of ER-ACE, which uses its standard cosine classifier, over ER from 12.8 to 0.3 pp, within the equivalence margin. DER and DER++ retain 4.8 and 5.7 pp and are also more accurate in group-restricted accuracy (Figure~\ref{fig:scope}a; Appendix~\ref{app:cal}). Among the tested methods, only logit replay keeps a large corrected advantage over ER, so as-trained gains do not reveal how much of a method's advantage remains after correction.

\paragraph{Distillation objectives.}
Students learn CIFAR-100 from 50 labeled images per class and distill from a teacher trained on half the classes (E0) or additionally for one epoch on all classes (E1). The teacher-to-student KL divergence decomposes exactly into block-mass matching and teacher-mass-weighted within-block matching. We add either component, both components, or neither to the label loss, using eight teachers and offsets fitted on held-out training images. Mass matching costs 9.7 pp at E0 as trained but is within $\pm1$ pp of label-only training after the offset, whereas within-block matching keeps its gains at both stages (Figure~\ref{fig:scope}b). As on $H$, a separate E1 cohort reassigns non-gold target probabilities, preserving gold probability, block mass, and entropy. After the offset, the original target exceeds the reassigned one by 8.1 pp, a flat within-block target by 5.7, and a class template by 3.4 (Figure~\ref{fig:scope}c; Appendix~\ref{app:kd-content}). The corrected gain thus depends on non-gold class correspondence.

\section{Conclusion}

A large deletion cost shows that replay matching matters, not what its target must preserve. Separating the matching constraint, the stored values, and class correspondence, and evaluating each model before and after a task-level offset, reveals two requirements of different kinds in the tested fixed-head setting: suitable constants can replace the unrefreshed scores of later-learned classes, and most of the cost of deleting their matching is correctable, whereas the smaller cost of reassigning known-class scores persists, both in replay and in image distillation. Our evidence covers offline training on two image datasets and three ResNet configurations; both the fixed anchor and the offsets, which assume a known task share, are diagnostics rather than methods. In a text check, the within-block distillation gain depends on the class partition (Appendix~\ref{app:kd-text}). Within this scope, replay studies should pair deletion with replacement and correspondence controls, and report accuracy before and after bias correction.
\label{maintext:end}
\section*{AI Use Statement}
Generative AI tools were used in this work to assist with language editing, as well as with the preparation and refinement of figures and tables. All AI-assisted materials were carefully reviewed and revised by the authors to ensure their accuracy and faithful representation of the research. These tools were not used to generate experimental results, perform scientific analyses, or formulate research conclusions. The authors retain full responsibility for all content presented in this paper.

\section*{Reproducibility Statement}
Sections~\ref{sec:v6-design} and~\ref{sec:v76-experiments} and Appendix~\ref{app:v6-robustness} give the training settings, target constructions, and statistical procedures, and Appendix~\ref{app:cal} gives the task-level offset readout. The anonymized supplementary material contains the training code for every intervention, replay method, and distillation student, the configuration of each cohort, the per-seed final accuracies, diagnostic readouts, and offset readouts, and scripts that recompute every main-text number, the settings table, the offset tables, and every figure from those records, checking each printed number against its record; the other appendix tables are paired statistics over the same records. Checkpoints are not included; each offset record names the checkpoint it was computed from. All datasets are public. The supplementary code, experimental records, and reproduction scripts are included as an ancillary archive with this arXiv version; an anonymized copy is also available at \url{https://anonymous.4open.science/r/replay-preserve-E22B/}.
\bibliographystyle{ces_linked}
\bibliography{refs,refs_v6,refs_v6_1}
\clearpage
\appendix
\setcounter{figure}{0}\setcounter{table}{0}
\renewcommand{\thefigure}{S\arabic{figure}}
\renewcommand{\thetable}{S\arabic{table}}
\renewcommand{\theHfigure}{supp.\arabic{figure}}
\renewcommand{\theHtable}{supp.\arabic{table}}
\begin{center}
{\Large\bfseries Supplementary Material}\\[3pt]
\end{center}

\paragraph{Protocol and notation.} Each stored logit vector is the output on the augmented training view in the forward pass that computes the loss, before that step's parameter update, and is stored with the non-augmented image; memory is filled by reservoir sampling over the stream. Task-selection accuracy $A_{\rm task}$ assigns an example to the task with the largest summed softmax probability \citep{kim2022theoretical}. Differences are paired by seed; intervals are paired $t$ intervals over seeds, equivalence uses two one-sided $t$ tests at the stated margin, and prespecified families are Holm-corrected. Table labels map to the main text as follows: Niso is Record mean; Iso $H$ (Iso, Mean $H$) is $H$ block mean; Perm $H$ (Permuted) is reassigned $H$; Self is Record mean in the pooled-reference implementation; in the native-$N$ reassignment cohort, Original is Full and Mean $N$ is its mean-valued arm. $A_{\rm task}$ (defined above) and $C_{\rm task}$ (Section~\ref{sec:replay-evaluation}) are the task-selection and within-task readouts; $A$ and $C$ are the same readouts with the 20 CIFAR-100 superclasses in place of the ten task blocks. The pooled-reference policies (Pooled, Self, Shuffle) and $H$ block mean are implemented as separate model classes that share the DER++ replay loop with the main class and differ only in the target edit; contrasts across classes are marked and read descriptively.

\section{\texorpdfstring{$N$}{N}-Target Replacement and Retained Constraints}\label{app:v6-anatomy}\label{app:v6-specificity-results}\label{app:v6-n}

\subsection{Replacement and scalar-anchor comparisons}\label{app:v6-numbers}
\begin{table}[H]
\centering\small
\caption{\textbf{Simplified $N$ targets retain accuracy, whereas deletion is costly.} Terminal class-incremental accuracy on CIFAR-100 with ResNet-18; twenty paired units. Untreated supports and label losses remain supervised.}
\label{tab:s41-replacement}
\begin{tabular}{@{}llr@{}}\toprule
Policy & Reference on $N$ & Accuracy (\%)\\\midrule
Full & Cached target vector & 37.62\\
Record mean (Niso) & Record-specific scalar & 37.30\\
Generic & Independently calibrated anchor & 37.09\\
Delete $N$ & Matching term removed & 22.46\\\bottomrule
\end{tabular}
\par\medskip
\begin{minipage}{.96\linewidth}\footnotesize
Figure~\ref{fig:s41-evidence} reports paired equivalence intervals; similar rounded arm means alone do not establish equivalence.
Generic remains teacher-calibrated.
Fixed-anchor tests use the first eight and the remaining twelve of these units as two separate sets (Table~\ref{tab:v77-fixed-v6}).
\end{minipage}
\end{table}

\begin{table}[H]
\centering\small
\caption{\textbf{Retaining the constraint benefits older tasks at a latest-task cost.} Paired accuracy differences in pp with 95\% intervals. Rows use separate cohorts; the component studies use Generic on CIFAR-100 and record mean on TinyImageNet.}
\label{tab:v75-population}
\begin{tabular}{@{}llcc@{}}\toprule
Dataset ($n$) & Contrast & Older tasks & Latest task \\\midrule
CIFAR-100 (12) & Full $-$ Delete $N$ & $+16.168$ & $-6.617$ \\
 & & $[15.150,17.185]$ & $[-8.492,-4.741]$ \\
CIFAR-100 (20) & Full $-$ Delete $N$ & $+17.832$ & $-8.855$ \\
 & & $[17.034,18.630]$ & $[-10.765,-6.945]$ \\
CIFAR-100 (20) & Full $-$ Perm $H$ & $+5.447$ & $-1.925$ \\
 & & $[4.815,6.080]$ & $[-3.146,-0.704]$ \\
CIFAR-100 (20) & Both $-$ Neither & $+17.101$ & $-7.650$ \\
 & & $[16.343,17.858]$ & $[-9.139,-6.161]$ \\
TinyImageNet (8) & Both $-$ Neither & $+10.532$ & $-7.588$ \\
 & & $[9.242,11.822]$ & $[-9.875,-5.300]$ \\\bottomrule
\end{tabular}
\end{table}

A terminal-memory audit of twelve Full and twelve Pooled runs of the pooled-reference cohort gives, as the average over runs of each run's median record, a root-mean-square residual of 0.344 logits on the stored $N$ scores and 1.884 on the stored non-gold $H$ scores; these terminal measurements do not describe the whole training trajectory. Full--Niso meets the declared equivalence margin on all five accuracy readouts in Table~\ref{tab:app-principal-replacement}. Generic--Niso is the prespecified primary Generic contrast. The post-hoc Generic--Full deployment difference is $-0.538$ pp (90\% CI $[-0.901,-0.175]$; 95\% CI $[-0.977,-0.099]$; TOST $p=.020$). Its task-selection interval $[-1.060,-0.204]$ crosses the equivalence boundary ($p=.077$).

Within each minibatch and insertion-task stratum, Pooled replaces record means by their stratum mean, whereas Shuffle cyclically reassigns those means; singleton strata retain their own reference. The pooled-reference cohort tested these policies on twelve fresh units after a four-unit exploration. Its primary Pooled--Full difference is $-0.220$ pp (90\% CI $[-0.704,0.264]$; TOST $p=.00728$). The post-hoc Pooled--Self and Self--Full intervals are $[-0.732,0.506]$ and $[-0.585,0.372]$; these linked contrasts share policies and do not add independent evidence. Pooled--Full also meets the secondary equivalence margins on older and latest tasks ($p=.0196$ and $.0156$). Because Full/Delete and Pooled/Self/Shuffle use separate model classes sharing one replay loop, Pooled--Full crosses implementations; Pooled--Self changes only the anchor within one implementation. Stratum occupancy was not recorded, so these comparisons do not quantify how often pooling or shuffling changed an individual reference.

In the prespecified two-member $C_{\rm task}$ family, Self--Shuffle meets the equivalence margin ($p=4.770\times10^{-5}$; Holm-adjusted $9.54\times10^{-5}$), while Shuffle--Pooled is unresolved (two-sided difference-test $p=.41682$; Table~\ref{tab:app-principal-n}). Self--Shuffle also meets the margin on aggregate $J$ in a descriptive test ($p=.0222$), but not on task selection (90\% CI $[-1.052,-0.023]$) or older/latest populations ($p=.0777/.2324$). Thus the observed tolerance of shuffled scalar references depends on the endpoint.

{\small\setlength{\tabcolsep}{4pt}
\begin{longtable}{@{}llrrr@{}}
\caption{Record-mean replacement: all accuracy endpoints (20 paired units). Differences in pp, with paired 95\% intervals.}\label{tab:app-principal-replacement}\\
\toprule
Comparison & Endpoint & Mean & 95\% interval & Positive/$n$\\\midrule\endfirsthead
\toprule
Comparison & Endpoint & Mean & 95\% interval & Positive/$n$\\\midrule\endhead
Full -- Niso & J & +0.329 & $[-0.056,+0.715]$ & 12/20\\
Full -- Niso & $A_{\rm task}$ & +0.236 & $[-0.221,+0.694]$ & 12/20\\
Full -- Niso & $C_{\rm task}$ & +0.092 & $[-0.190,+0.374]$ & 10/20\\
Full -- Niso & A & -0.009 & $[-0.457,+0.439]$ & 9/20\\
Full -- Niso & C & +0.246 & $[-0.024,+0.516]$ & 12/20\\
Niso -- Delete N & J & +14.836 & $[+14.212,+15.460]$ & 20/20\\
Niso -- Delete N & $A_{\rm task}$ & +16.590 & $[+15.830,+17.350]$ & 20/20\\
Niso -- Delete N & $C_{\rm task}$ & -4.802 & $[-5.210,-4.394]$ & 0/20\\
Niso -- Delete N & A & +14.950 & $[+14.209,+15.691]$ & 20/20\\
Niso -- Delete N & C & +5.652 & $[+4.939,+6.365]$ & 20/20\\
\bottomrule\end{longtable}}

{\small\setlength{\tabcolsep}{4pt}
\begin{longtable}{@{}llrrr@{}}
\caption{Deletion and anchor interventions: task-coordinate endpoints (12 paired units). Differences in pp, with paired 95\% intervals. Only Pooled--Full ($J$, $\pm1$ pp TOST) and the two-member $C_{\rm task}$ family were prespecified. The diagnostic Full--Delete $J$ value here is 13.886 pp, versus 13.889 pp in the frozen training endpoint.}\label{tab:app-principal-n}\\
\toprule
Comparison & Endpoint & Mean & 95\% interval & Positive/$n$\\\midrule\endfirsthead
\toprule
Comparison & Endpoint & Mean & 95\% interval & Positive/$n$\\\midrule\endhead
Full -- Delete N & J & +13.886 & $[+13.107,+14.665]$ & 12/12\\
Full -- Delete N & $A_{\rm task}$ & +15.750 & $[+14.855,+16.645]$ & 12/12\\
Full -- Delete N & $C_{\rm task}$ & -5.116 & $[-5.654,-4.578]$ & 0/12\\
Pooled -- Full & J & -0.220 & $[-0.813,+0.373]$ & 3/12\\
Pooled -- Full & $A_{\rm task}$ & -0.348 & $[-0.985,+0.288]$ & 3/12\\
Pooled -- Full & $C_{\rm task}$ & -0.058 & $[-0.572,+0.455]$ & 6/12\\
Pooled -- Self (post hoc) & J & -0.113 & $[-0.872,+0.645]$ & 5/12\\
Self -- Full (post hoc) & J & -0.107 & $[-0.693,+0.479]$ & 6/12\\
Self -- Shuffle & J & -0.388 & $[-0.981,+0.206]$ & 4/12\\
Self -- Shuffle & $A_{\rm task}$ & -0.538 & $[-1.168,+0.093]$ & 5/12\\
Self -- Shuffle & $C_{\rm task}$ & -0.037 & $[-0.393,+0.318]$ & 4/12\\
Shuffle -- Pooled & J & +0.501 & $[-0.131,+1.133]$ & 9/12\\
Shuffle -- Pooled & $A_{\rm task}$ & +0.673 & $[-0.027,+1.373]$ & 10/12\\
Shuffle -- Pooled & $C_{\rm task}$ & +0.163 & $[-0.261,+0.586]$ & 8/12\\
\bottomrule\end{longtable}}

{\small\setlength{\tabcolsep}{4pt}
\begin{longtable}{@{}llrrr@{}}
\caption{Anchor interventions evaluated on older and newest tasks, recovered from the same sixty terminal per-task vectors (12 paired units). Differences in pp, with paired 95\% intervals. Full--Delete population contrasts are in Table~\ref{tab:v75-population}.}\label{tab:app-principal-pop}\\
\toprule
Comparison & Endpoint & Mean & 95\% interval & Positive/$n$\\\midrule\endfirsthead
\toprule
Comparison & Endpoint & Mean & 95\% interval & Positive/$n$\\\midrule\endhead
Pooled -- Full & old & -0.249 & $[-0.955,+0.457]$ & 3/12\\
Pooled -- Full & latest & -0.025 & $[-0.894,+0.844]$ & 7/12\\
Self -- Shuffle & old & -0.484 & $[-1.228,+0.260]$ & 5/12\\
Self -- Shuffle & latest & +0.517 & $[-0.888,+1.922]$ & 7/12\\
Shuffle -- Pooled & old & +0.528 & $[-0.257,+1.312]$ & 9/12\\
Shuffle -- Pooled & latest & +0.250 & $[-0.708,+1.208]$ & 7/12\\
\bottomrule\end{longtable}}

\paragraph{Generic anchor.} Generic sits between Record mean and a fixed anchor: it uses $a_{s,t}(j)=m_{s,t}+\sigma_{s,t}\xi(j)$ on $N$, where $m_{s,t}$ and $\sigma_{s,t}$ are the mean and sample standard deviation of record means for insertion task $s$ and replay task $t$ in a separate DER++ run, and $\xi(j)$ is a fixed pseudo-random offset of reservoir slot $j$. It removes the record's own mean but keeps a teacher calibration. On CIFAR-100 it reaches 37.1\% and is equivalent to Record mean ($-0.21$ pp, 90\% CI $[-0.68,0.26]$) and, post hoc, to Full (above).

\subsection{Fixed-anchor replacements}\label{app:v77-fixed}
Table~\ref{tab:v77-fixed-v6} compares each fixed $N$ anchor with Full on the same seeds. The initial $a=-2/0$ family and later $a=-1/-4$ family are corrected separately. On these eight units, $a=-2$, $-1$, and $-4$ meet the $\pm1$ pp equivalence criterion after their within-family Holm corrections. On the remaining twelve units of the main cohort, specified before those runs, $a=-2$ and $a=-4$ again meet it, whereas $a=-1$ does not ($+0.56$ pp, 90\% CI $[0.06,1.06]$) and $a=0$ again does not establish it; the constant set reported in the main text is therefore $\{-2,-4\}$. Over all twenty units, the descriptive costs are $+0.72$, $+0.41$, $+0.23$, and $+0.00$ pp for $a=0,-1,-2,-4$. These variants retain replay labels and the $H/F$ matching targets.
\begin{table}[H]\centering\footnotesize\setlength{\tabcolsep}{3pt}
\caption{Fixed $N$ references retain the tested benefit for selected constants. Accuracy and Full-minus-variant cost are in percentage points. Intervals are paired 90\% intervals for $\pm1$ pp equivalence; Holm $p$ is adjusted within the specified family, not across families. Old/new costs are descriptive; --- where only the training-log endpoint exists. Tiny $a=-2$ and memory rows use their own Full controls. Families E--H were specified before their runs; E and ``one'' are single tests.}\label{tab:v77-fixed-v6}
\resizebox{\linewidth}{!}{\begin{tabular}{@{}llrrrrlll@{}}\toprule
Method & Anchor & $n$ & Acc. & Cost & 90\% CI & Family & Holm $p$ & Old/new cost \\\midrule
DER++ & $a=-1$ & 8 & 37.7 & $+0.19$ & $[-0.10,+0.47]$ & later B & .001 & $+0.36/-1.40$\\
DER++ & $a=-2$ & 8 & 37.5 & $+0.44$ & $[+0.04,+0.85]$ & initial A & .035 & $+0.52/-0.26$\\
DER++ & $a=-4$ & 8 & 37.9 & $+0.03$ & $[-0.58,+0.64]$ & later B & .010 & $-0.08/+1.01$\\
DER++ & $a=0$ & 8 & 37.1 & $+0.80$ & $[+0.39,+1.21]$ & initial A & .189 & $+1.07/-1.61$\\
DER++ & $N/F$ fixed, $a=-2$ & 8 & 37.1 & $+0.83$ & $[+0.40,+1.25]$ & C & .234 & $+1.15/-2.05$\\
DER++, remaining 12 & $a=-2$ & 12 & 37.4 & $+0.08$ & $[-0.71,+0.87]$ & E & .031 & ---\\
DER++, remaining 12 & $a=-1$ & 12 & 36.9 & $+0.56$ & $[+0.06,+1.06]$ & F & .071 & ---\\
DER++, remaining 12 & $a=-4$ & 12 & 37.5 & $-0.02$ & $[-0.79,+0.75]$ & F & .044 & ---\\
DER++, remaining 12 & $a=0$ & 12 & 36.8 & $+0.66$ & $[+0.26,+1.06]$ & one & .080 & ---\\
DER & $a=-2$ & 12 & 34.6 & $+0.16$ & $[-0.40,+0.72]$ & D & .010 & $+0.08/+0.83$\\
DER & $a=-1$ & 12 & 34.7 & $+0.05$ & $[-0.44,+0.53]$ & G & .005 & ---\\
DER & $a=-4$ & 12 & 35.1 & $-0.30$ & $[-0.82,+0.21]$ & G & .017 & ---\\
Tiny DER++ & $a=-2$ & 8 & 19.8 & $-0.65$ & $[-1.48,+0.18]$ & one & .227 & ---\\
DER++, mem.\ 200 & $a=-2$ & 8 & 24.4 & $-0.28$ & $[-0.89,+0.33]$ & H & .030 & ---\\
DER++, mem.\ 2,000 & $a=-2$ & 8 & 52.8 & $+0.12$ & $[-0.25,+0.48]$ & H & .002 & ---\\\bottomrule
\end{tabular}}
\end{table}

\subsection{Component interactions beyond deployment}\label{app:v75-components}\label{app:v7-generic-factorial}
\begin{table}[H]
\centering\small
\caption{\textbf{Both retained components outperform either alone.} Accuracy (\%) under fixed coefficients; bold marks the highest observed mean among the four component policies, not a significance test. Paired contrasts below use pp and 95\% intervals.}
\label{tab:s41-components}\label{tab:v76-components}
\begin{tabular}{@{}lccrr@{}}\toprule
& \multicolumn{2}{c}{Retained terms} & CIFAR-100 & TinyImageNet\\
Policy & Level & Dispersion & Generic, $n=20$ & Record mean, $n=8$\\\midrule
Neither & -- & -- & 22.46 & 10.73\\
Level only & $\checkmark$ & -- & 24.78 & 11.67\\
Dispersion only & -- & $\checkmark$ & 28.98 & 12.06\\
Both & $\checkmark$ & $\checkmark$ & \textbf{37.09} & \textbf{19.45}\\\midrule
\multicolumn{3}{l}{Both $-$ Level only} &
\shortstack{$12.304$\\$[11.661,12.946]$} &
\shortstack{$7.780$\\$[6.803,8.757]$}\\
\multicolumn{3}{l}{Both $-$ Dispersion only} &
\shortstack{$8.103$\\$[7.445,8.761]$} &
\shortstack{$7.385$\\$[6.401,8.369]$}\\
\multicolumn{3}{l}{Interaction $I_J$} &
\shortstack{$5.781$\\$[5.137,6.425]$} &
\shortstack{$6.445$\\$[5.454,7.436]$}\\\bottomrule
\end{tabular}
\par\medskip
\begin{minipage}{.96\linewidth}\footnotesize
$I_J=J_{\rm Both}-J_{\rm Level}-J_{\rm Dispersion}+J_{\rm Neither}$.
The CIFAR-100 interaction is descriptive; the TinyImageNet deployment interaction was prespecified.
The two datasets use different anchors and separate cohorts.
\end{minipage}
\end{table}
The Both--Level--Dispersion+Neither interaction is positive on older tasks and negative on the latest task in both datasets. These descriptive population readouts show that the aggregate interaction does not benefit both populations equally.
\begin{table}[H]\centering\small
\caption{\textbf{Level and dispersion interact positively on older tasks and negatively on the latest task.} Differences in pp with paired 95\% intervals.}\label{tab:app-pop-interaction}
\begin{tabular}{@{}llrr@{}}\toprule
Dataset & Endpoint & Interaction & 95\% CI\\\midrule
CIFAR-100 & Older & $+7.179$ & $[6.378,7.980]$\\
CIFAR-100 & Latest & $-6.800$ & $[-8.233,-5.367]$\\
TinyImageNet & Older & $+7.843$ & $[6.547,9.139]$\\
TinyImageNet & Latest & $-6.138$ & $[-8.795,-3.480]$\\\bottomrule
\end{tabular}
\end{table}

\subsection{Factorial diagnostics and doubled coefficients}\label{app:v77-double}
Table~\ref{tab:v77-factorial-v6} measures interactions among trained-policy outcomes; the matching-loss components remain additive. The CIFAR-100 factorial uses Generic, while TinyImageNet uses the record mean in a separate cohort. Table~\ref{tab:v77-double-v6} compares the joint policy with each single term at twice its original coefficient.
\begin{table}[H]\centering\footnotesize\setlength{\tabcolsep}{3pt}
\caption{Level--dispersion interactions on accuracy and exact loss terms (mean and paired 95\% interval). Positive loss interaction denotes a larger reduction by the joint policy.}\label{tab:v77-factorial-v6}
\resizebox{\linewidth}{!}{\begin{tabular}{@{}lrllll@{}}\toprule
Setting & $n$ & $I_J$ (pp) & $I_{C_{\rm task}}$ (pp) & $I_{L_{\rm task}}$ (nats) & $I_{L_{\rm within}}$ (nats)\\\midrule
CIFAR-100, Generic & 20 & $+5.78\ [5.14,6.43]$ & $-1.45\ [-1.78,-1.12]$ & $+0.64\ [0.58,0.69]$ & $-0.02\ [-0.03,-0.01]$\\
TinyImageNet, record mean & 8 & $+6.44\ [5.45,7.44]$ & $+1.15\ [0.48,1.81]$ & $+1.41\ [1.25,1.57]$ & $+0.09\ [0.06,0.13]$\\\bottomrule
\end{tabular}}
\end{table}
\begin{table}[H]\centering\footnotesize\setlength{\tabcolsep}{4pt}
\caption{Doubling a single term does not recover the joint policy on CIFAR-100 DER++ ($n=8$). Both uses two terms at the original coefficient; each single-term policy uses twice the original coefficient. Differences are paired means with 95\% intervals; loss columns are reductions by Both, so positive values mean lower loss under Both.}\label{tab:v77-double-v6}
\begin{tabular}{@{}llll@{}}\toprule
Comparison & $J$ (pp) & $L_{\rm task}$ (nats) & $L_{\rm within}$ (nats)\\\midrule
Both $-$ doubled level & $+12.49\ [11.43,13.55]$ & $+1.59\ [1.46,1.71]$ & $-0.17\ [-0.19,-0.15]$\\
Both $-$ doubled dispersion & $+7.32\ [5.76,8.87]$ & $+0.75\ [0.60,0.90]$ & $+0.02\ [0.01,0.04]$\\\bottomrule
\end{tabular}
\end{table}

\section{Class Correspondence and Prediction Behavior}\label{app:v75-correspondence}\label{app:v75-correspondence-readouts}\label{app:v7-energy}
\subsection{Reassignment on historical and later classes}
Full--Permuted $H$ changes both task-selection and within-task accuracy, with the larger difference on within-task accuracy; by contrast, native-$N$ reassignment meets the $\pm1$ pp equivalence criterion on deployment accuracy. The following tables give the paired endpoint estimates behind this distinction.
{\small\setlength{\tabcolsep}{4pt}
\begin{longtable}{@{}llrrr@{}}
\caption{Historical-class reassignment readouts and the three-arm deployment comparison (20 paired units). Differences in pp, with paired 95\% intervals. Starred contrasts cross implementations and are descriptive.}\label{tab:app-principal-h}\\
\toprule
Comparison & Endpoint & Mean & 95\% interval & Positive/$n$\\\midrule\endfirsthead
\toprule
Comparison & Endpoint & Mean & 95\% interval & Positive/$n$\\\midrule\endhead
Full -- Perm H & J & +4.714 & $[+4.213,+5.214]$ & 20/20\\
Full -- Perm H & $A_{\rm task}$ & +4.689 & $[+4.172,+5.206]$ & 20/20\\
Full -- Perm H & $C_{\rm task}$ & +12.771 & $[+12.288,+13.255]$ & 20/20\\
Full -- Perm H & A & +5.098 & $[+4.517,+5.678]$ & 20/20\\
Full -- Perm H & C & +5.517 & $[+5.111,+5.923]$ & 20/20\\
Full -- Iso H* & J & +3.405 & $[+3.036,+3.774]$ & 20/20\\
Iso -- Perm H* & J & +1.309 & $[+0.918,+1.699]$ & 19/20\\
\bottomrule\end{longtable}}

For native-$N$ reassignment, Original--Permute passes TOST at $\pm1$ pp ($p=.0020$, Holm-adjusted $.0041$); Original--Mean passes at $\pm1.5$ pp ($p=.00007$). The twelve-unit deployment endpoint comes from frozen training records; task diagnostics were computed separately (Table~\ref{tab:app-n-permutation}).
{\footnotesize\setlength{\tabcolsep}{3pt}\begin{longtable}{@{}llrrrl@{}}
\caption{Coordinate reassignment on N: frozen primary and separately identified secondary readouts.}\label{tab:app-n-permutation}\\
\toprule Comparison & Endpoint & Mean & 90\% CI & 95\% CI & Role\\\midrule\endfirsthead
\toprule Comparison & Endpoint & Mean & 90\% CI & 95\% CI & Role\\\midrule\endhead
Original -- Permute N & J & -0.011 & $[-0.502,+0.481]$ & $[-0.613,+0.592]$ & primary, $\pm1$ pp\\
Original -- Mean N & J & +0.222 & $[-0.181,+0.624]$ & $[-0.272,+0.715]$ & secondary, $\pm1.5$ pp\\
Perm. -- Orig. N & old & +0.088 & $[-0.490,+0.666]$ & $[-0.621,+0.797]$ & descr. population\\
Perm. -- Orig. N & latest & -0.683 & $[-1.470,+0.103]$ & $[-1.647,+0.281]$ & descr. population\\
Orig. -- Perm. N (instr.) & $A_{\rm task}$ & -0.294 & $[-0.831,+0.243]$ & $[-0.952,+0.364]$ & instr. projection\\
Orig. -- Perm. N (instr.) & $C_{\rm task}$ & +0.013 & $[-0.220,+0.247]$ & $[-0.273,+0.300]$ & instr. projection\\
\bottomrule\end{longtable}}

\subsection{Restricted historical-class reassignment}\label{app:v77-h-blocks}
The restrictions in Table~\ref{tab:v77-h-blocks-v6} edit different amounts of the $H$ target. Their costs cannot be added: records stored during the first task have no other previously seen block.
\begin{table}[H]\centering\small
\caption{Both the record's own task block and other seen blocks carry $H$ correspondence. CIFAR-100 DER++; the first eight units and, specified before their runs, the remaining twelve units of the main cohort. Full-minus-variant cost in pp with 95\% intervals; one-sided replication tests Holm-adjusted over the two restrictions ($p<10^{-4}$ each).}\label{tab:v77-h-blocks-v6}
\begin{tabular}{@{}llrrr@{}}\toprule
Reassignment restricted to & Units & Cost & 95\% CI & Positive/$n$\\\midrule
Own task block only & first 8 & $+1.79$ & $[1.08,2.49]$ & 8/8\\
Own task block only & remaining 12 & $+1.76$ & $[1.09,2.42]$ & 11/12\\
Other seen task blocks only & first 8 & $+3.72$ & $[2.66,4.79]$ & 8/8\\
Other seen task blocks only & remaining 12 & $+3.86$ & $[3.11,4.61]$ & 12/12\\\bottomrule
\end{tabular}
\end{table}

\subsection{Residual-energy controls}
For an eligible coordinate set $B$ of a record, define per-coordinate centered energy $e_B=\|q_{i,B}-\mu_{q_i,B}\mathbf1\|_2^2/|B|$; $H$ excludes the ground-truth coordinate, while $N$ contains no ground-truth coordinate. We scale the selected centered values by $\sqrt{e_{\rm target}/e_{\rm native}}$ before applying either assignment. The $H$ comparisons use target energies $e_H$, $\sqrt{e_He_N}$, and $e_N$; \HNMcompanion. Records with fewer than two eligible coordinates on either support, or with $e_H$ or $e_N\leq10^{-12}$, pass through unchanged. At each energy, the original-versus-permuted comparison preserves the eligible target-value multiset, while leaving ground-truth and untreated coordinates unchanged.

\begin{figure}[H]
\centering
\includegraphics[width=.78\linewidth]{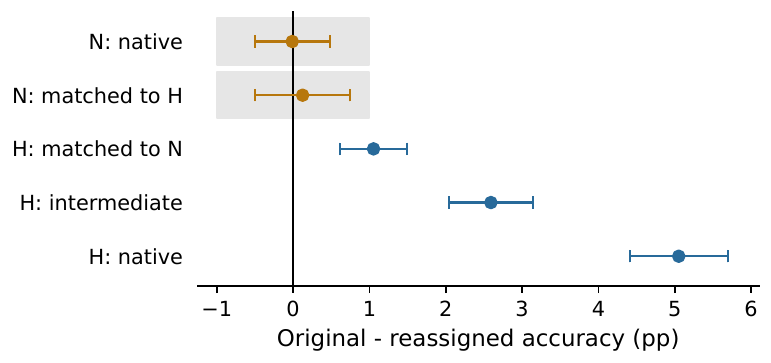}
\caption{\textbf{$H$ assignment remains costly after reducing its residual energy, while raised-energy $N$ reassignment remains equivalent.} Original-minus-reassigned accuracy on CIFAR-100, with twelve paired units per comparison. The $H$ rows and ``$N$: matched to $H$'' use within-task-block reassignment in one cohort; ``$N$: native'' is the across-$N$ shift, from a separate cohort. $H$ intervals are 95\%; $N$ intervals are 90\%, with shading indicating the $\pm1$ pp equivalence band.}
\label{fig:s42-energy}
\end{figure}

Figure~\ref{fig:s42-energy} shows that reassignment on $H$ remains costly as residual energy falls, whereas raised-energy $N$ reassignment remains within the one-point equivalence band.

The native-minus-$N$-matched $H$ cost is $+3.998$ pp (95\% CI $[3.247,4.750]$); native minus intermediate is $+2.460$ pp ($[1.919,3.001]$) and intermediate minus $N$-matched $+1.538$ pp ($[0.749,2.328]$). At native, intermediate, and low $H$ energy, the older/latest effects are $+5.931/-2.858$, $+2.988/-0.967$, and $+1.171/-0.000$ pp. Raised-energy $N$ changes older-task accuracy by $+0.151$ pp ($[-0.764,1.066]$) and latest-task accuracy by $-0.108$ pp ($[-1.359,1.142]$), with paired 95\% intervals.

\subsection{Exact task/within-task loss split}\label{app:v77-loss-split}
Let $p_j$ be the softmax probability over seen classes and $T(y)$ the classes in the true task of label $y$. For each example, cross-entropy splits exactly into a task term and a within-task term,
\begin{equation}
 -\log p_y
 =
 \underbrace{-\log\sum_{j\in T(y)}p_j}_{L_{\rm task}}
 +
 \underbrace{-\log\frac{p_y}{\sum_{j\in T(y)}p_j}}_{L_{\rm within}}.
 \label{eq:v6-task-within-ce}
\end{equation}
The first term measures probability assigned to the true task; the second measures discrimination among its classes, and the task-level offset leaves it exactly unchanged. Table~\ref{tab:v77-loss-split-v6} applies this split to one re-evaluation of retained checkpoints. Rows sharing seeds or Full controls are dependent. The accuracy contrasts in this re-evaluation differ from frozen training endpoints by at most .0075 pp. The native-$N$ reassignment cohort is excluded because this evaluation covers only three of its twelve Full checkpoints. Across the nine listed $H$ comparisons, the within-task term contributes 61\%--85\% of the mean reduction in total cross-entropy, calculated as a ratio of mean loss reductions.
{\footnotesize\setlength{\tabcolsep}{4pt}\renewcommand{\arraystretch}{0.88}
\begin{longtable}{@{}lrlll@{}}
\caption{Exact cross-entropy split: first-policy accuracy gain $J$ (pp) and loss reductions (nats), with paired 95\% intervals. C100 denotes CIFAR-100; ``res.'' denotes residual energy.}\label{tab:v77-loss-split-v6}\\
\toprule
Comparison & $n$ & $J$ (pp) & $L_{\rm task}$ reduction & $L_{\rm within}$ reduction\\\midrule
\endfirsthead
\toprule
Comparison & $n$ & $J$ (pp) & $L_{\rm task}$ reduction & $L_{\rm within}$ reduction\\\midrule
\endhead
\multicolumn{5}{@{}l}{\emph{Full minus Delete $N$}}\\
C100 DER++ & 20 & $+15.16$ & $+1.77\ [1.69,1.85]$ & $-0.17\ [-0.18,-0.17]$\\
C100 DER++, second cohort & 12 & $+13.89$ & $+1.67\ [1.60,1.74]$ & $-0.19\ [-0.20,-0.17]$\\
C100 DER & 12 & $+23.74$ & $+17.86\ [17.32,18.41]$ & $-0.27\ [-0.28,-0.25]$\\
TinyImageNet DER++ & 8 & $+8.41$ & $+2.77\ [2.67,2.87]$ & $-0.31\ [-0.34,-0.28]$\\\midrule
\multicolumn{5}{@{}l}{\emph{Pooled reference minus Delete $N$}}\\
C100, memory 200 & 8 & $+9.47$ & $+2.22\ [2.13,2.32]$ & $-0.18\ [-0.20,-0.17]$\\
C100, memory 500 & 8 & $+14.54$ & $+1.64\ [1.54,1.74]$ & $-0.18\ [-0.19,-0.17]$\\
C100, memory 2000 & 8 & $+11.86$ & $+0.78\ [0.74,0.82]$ & $-0.17\ [-0.18,-0.15]$\\\midrule
\multicolumn{5}{@{}l}{\emph{Full minus reassigned $H$}}\\
C100 DER++ & 20 & $+4.71$ & $+0.13\ [0.10,0.17]$ & $+0.39\ [0.38,0.41]$\\
C100 DER++, native res. & 12 & $+5.05$ & $+0.21\ [0.16,0.26]$ & $+0.38\ [0.37,0.39]$\\
C100 DER++, intermediate res. & 12 & $+2.59$ & $+0.11\ [0.07,0.14]$ & $+0.16\ [0.15,0.18]$\\
C100 DER++, $N$-size res. & 12 & $+1.05$ & $+0.04\ [0.01,0.08]$ & $+0.07\ [0.06,0.08]$\\
C100 DER & 8 & $+4.56$ & $+0.10\ [0.00,0.20]$ & $+0.43\ [0.40,0.46]$\\
TinyImageNet DER++ & 4 & $+6.09$ & $+0.26\ [0.04,0.49]$ & $+0.69\ [0.66,0.72]$\\
C100 DER++, memory 200 & 8 & $+2.61$ & $+0.07\ [-0.13,0.27]$ & $+0.39\ [0.35,0.44]$\\
C100 DER++, memory 500 & 8 & $+3.75$ & $+0.07\ [0.04,0.11]$ & $+0.37\ [0.35,0.40]$\\
C100 DER++, memory 2000 & 8 & $+3.70$ & $+0.10\ [0.08,0.13]$ & $+0.30\ [0.27,0.32]$\\\midrule
\multicolumn{5}{@{}l}{\emph{Full minus replacement}}\\
C100 DER++, record mean & 20 & $+0.33$ & $+0.02\ [-0.01,0.05]$ & $+0.01\ [-0.00,0.01]$\\
C100 DER++, Generic & 20 & $+0.54$ & $+0.03\ [-0.00,0.06]$ & $+0.01\ [-0.00,0.01]$\\
C100 DER++, fixed $a=-2$ & 8 & $+0.44$ & $+0.04\ [0.01,0.08]$ & $-0.00\ [-0.01,0.01]$\\
C100 DER, record mean & 12 & $+0.13$ & $+0.01\ [-0.04,0.07]$ & $+0.00\ [-0.01,0.01]$\\\bottomrule
\end{longtable}}

\subsection{Terminal classifier-norm equalization}\label{app:v77-norm}
For each terminal checkpoint, we rescale classifier rows in every task block to the mean row norm over all seen classes, leaving biases fixed. Weight Aligning \citep{zhao2020maintaining} instead rescales new-class weights by an old-to-new norm ratio after each task; our terminal control is therefore a different intervention. Keeping biases can change within-task rankings, although the measured change in within-task accuracy is at most .06 pp across corrected checkpoints. Table~\ref{tab:v77-norm-v6} shows that all tested gaps remain; the DER deletion gap narrows from 23.74 to 15.05 pp.
\begin{table}[H]\centering\footnotesize\setlength{\tabcolsep}{3pt}
\caption{Terminal task-block row-norm equalization does not remove the tested gaps. Full-minus-edited accuracy gaps in pp before and after correction; the last column is the paired 95\% interval for the change in the gap.}\label{tab:v77-norm-v6}
\resizebox{\linewidth}{!}{\begin{tabular}{@{}llrrrl@{}}\toprule
Setting & Comparison & $n$ & Before & After & Change [95\% CI]\\\midrule
C100 DER++ & Full vs Delete $N$ & 20 & 15.16 & 15.67 & $+0.51\ [0.08,0.93]$\\
C100 DER & Full vs Delete $N$ & 12 & 23.74 & 15.05 & $-8.70\ [-9.33,-8.06]$\\
TinyImageNet DER++ & Full vs Delete $N$ & 8 & 8.41 & 6.60 & $-1.81\ [-2.00,-1.62]$\\
C100 DER++ & Full vs Perm $H$ & 20 & 4.71 & 4.77 & $+0.06\ [-0.08,0.20]$\\
C100 DER & Full vs Perm $H$ & 8 & 4.56 & 4.60 & $+0.04\ [-0.18,0.25]$\\
TinyImageNet DER++ & Full vs Perm $H$ & 4 & 6.09 & 5.49 & $-0.60\ [-0.92,-0.27]$\\\bottomrule
\end{tabular}}
\end{table}

\section{Effects Across Datasets, Replay Objectives, and Memory Budgets}\label{app:v6-robustness}\label{app:v6-hdir}\label{app:v6-later}
\paragraph{Optimization.} Under the principal recipe, current and replay batches contain 32 examples; SGD uses no momentum or weight decay; its learning rate starts at 0.03 in each task and drops by a factor of 0.2 at epochs 35 and 45 of 50.

\begin{table}[H]\centering\footnotesize\setlength{\tabcolsep}{4pt}
\caption{\textbf{Selected replacement, deletion, and reassignment effects extend beyond the canonical comparison.} Accuracy differences in pp with paired intervals at the stated level; $^\dagger$: 90\% interval within $\pm1$ pp. Rows may use separate seeds; all use ResNet-18 except the ResNet-32 rows. Mean/Perm~$H$ are the block-mean and reassignment edits of Section~\ref{sec:replay-correspondence-design}. }\label{tab:v8-settings}\label{tab:v76-transfer}\label{tab:v75-transfer}
\begin{tabular}{@{}lllrl@{}}\toprule
Setting & Objective & Contrast & $n$ & Difference [CI]\\\midrule
CIFAR-100 & DER & Full $-$ Delete $N$ & 12 & $23.7\ [22.9,24.6]_{95\%}$\\
CIFAR-100 & DER & Full $-$ Record mean & 12 & $0.13\ [-0.28,0.54]_{90\%}$$^\dagger$\\
CIFAR-100 & DER & Full $-$ Fixed $N$, $a=-1$ & 12 & $0.05\ [-0.44,0.53]_{90\%}$$^\dagger$\\
CIFAR-100 & DER & Full $-$ Fixed $N$, $a=-2$ & 12 & $0.16\ [-0.40,0.72]_{90\%}$$^\dagger$\\
CIFAR-100 & DER & Full $-$ Fixed $N$, $a=-4$ & 12 & $-0.30\ [-0.82,0.21]_{90\%}$$^\dagger$\\
CIFAR-100 & DER & Full $-$ Perm $H$ & 8 & $4.56\ [3.71,5.42]_{95\%}$\\
\addlinespace[2pt]
TinyImageNet & DER++ & Full $-$ Delete $N$ & 8 & $8.41\ [7.07,9.75]_{95\%}$\\
TinyImageNet & DER++ & Full $-$ Record mean & 8 & $-0.31\ [-1.31,0.69]_{90\%}$\\
TinyImageNet & DER++ & Full $-$ Fixed $N$, $a=-2$ & 8 & $-0.65\ [-1.48,0.18]_{90\%}$\\
TinyImageNet & DER++ & Full $-$ Mean $H$ & 4 & $4.05\ [3.20,4.91]_{95\%}$\\
TinyImageNet & DER++ & Full $-$ Perm $H$ & 4 & $6.09\ [5.11,7.07]_{95\%}$\\
TinyImageNet, 10 seeds & DER++ & Full $-$ Record mean & 10 & $0.29\ [-0.29,0.88]_{90\%}$$^\dagger$\\
TinyImageNet, 10 seeds & DER++ & Full $-$ Fixed $N$, $a=-2$ & 10 & $-0.03\ [-0.76,0.69]_{90\%}$$^\dagger$\\
\addlinespace[2pt]
CIFAR-100, ResNet-32 & DER++ & Full $-$ Delete $N$ & 20 & $6.28\ [5.79,6.77]_{95\%}$\\
CIFAR-100, ResNet-32 & DER++ & Full $-$ Fixed $N$, $a=-2$ & 20 & $-0.15\ [-0.60,0.29]_{90\%}$$^\dagger$\\
CIFAR-100, ResNet-32 & DER++ & Full $-$ Perm $H$ & 20 & $3.84\ [3.39,4.28]_{95\%}$\\
\addlinespace[2pt]
CIFAR-100, mem.\ 200 & DER++ & Full $-$ Delete $N$ & 8 & $9.13\ [8.06,10.2]_{95\%}$\\
CIFAR-100, mem.\ 200 & DER++ & Full $-$ Fixed $N$, $a=-2$ & 8 & $-0.28\ [-0.89,0.33]_{90\%}$$^\dagger$\\
CIFAR-100, mem.\ 200 & DER++ & Full $-$ Perm $H$ & 8 & $2.61\ [1.07,4.15]_{95\%}$\\
\addlinespace[2pt]
CIFAR-100, mem.\ 2{,}000 & DER++ & Full $-$ Delete $N$ & 8 & $11.7\ [11.0,12.5]_{95\%}$\\
CIFAR-100, mem.\ 2{,}000 & DER++ & Full $-$ Fixed $N$, $a=-2$ & 8 & $0.12\ [-0.25,0.48]_{90\%}$$^\dagger$\\
CIFAR-100, mem.\ 2{,}000 & DER++ & Full $-$ Perm $H$ & 8 & $3.70\ [3.14,4.26]_{95\%}$\\
\bottomrule\end{tabular}
\end{table}
\subsection{Dataset and replay-objective checks}
The seed-to-seed standard deviation of Full's final accuracy is 1.45 pp in the main CIFAR-100 cohort (20 seeds), 1.23 under DER (12), and 1.72 on TinyImageNet (8); it is lower in some smaller or larger-memory cohorts (1.04 on the first eight main-cohort seeds, 0.36 at memory 2,000). Our Full DER++ reaches 37.6\%; \citet{bosch2022xder} report 38.25\% for DER++ in the same ten-task CIFAR-100, memory-500 setting. TinyImageNet uses ten tasks of twenty classes, a preallocated 200-way ResNet-18, memory 500, 100 epochs per task, $\alpha=.2$, $\beta=.5$, and no learning-rate scheduler. The CIFAR-100 DER follow-up uses one matching-replay draw and no replay-label draw. Under DER, Full--Permuted changes older/latest accuracy by $+5.193/-1.100$ pp; the latest-task interval includes zero. On TinyImageNet, Full--Mean is $+4.053$ pp (95\% CI $[3.197,4.908]$), Mean--Permuted is $+2.035$ pp ($[1.021,3.049]$), and Full--Permuted changes older/latest accuracy by $+7.006/-2.175$ pp. These separate cohorts extend selected $H$ comparisons. A ResNet-32 cohort of 20 seeds (the principal recipe with only the backbone changed; Full reaches 23.5\%) tests a second architecture on Full, Delete $N$, the fixed anchor $a=-2$, and Perm $H$; its four prespecified tests pass under Holm correction (Table~\ref{tab:v8-settings}). A further TinyImageNet cohort compares Record mean and $a=-2$ with Full; it was specified with twelve seeds, of which ten ran (two did not finish before the evidence cutoff). Under the 90\% criterion used throughout the paper, both replacements meet the $\pm1$ pp margin (TOST $p=.027$ and $.019$). For this cohort we had also fixed, after one interim look, a stricter rule with 95\% intervals and Holm correction over the two comparisons; under that rule neither establishes equivalence. The earlier eight-seed TinyImageNet rows are unresolved at the same margin.

\subsection{Memory-capacity tests}\label{app:v6-scope-tables}
The correspondence and pooling sweeps use eight paired units at each memory size (200, 500, and 2,000) and share the Full policy within each size. Table~\ref{tab:v76-memory} gives pooling and deletion results; the following tables give correspondence, task, and population readouts. The two sweeps are analyzed separately; differences between frozen and diagnostic pipeline contrasts are at most .03 pp.

The prespecified Full$>$Iso$>$Permuted ordering on $J$ is unresolved at memory 500 because Iso--Permuted has two-sided $p=.270$ (Table~\ref{tab:app-h-budgets}). Full--Permuted remains positive at all three memory sizes; the $C_{\rm task}$ readouts favor both links but do not change the result of the specified $J$ test.

The pooling test specifies a $\pm1$ pp TOST and a positive Full--Delete control at each memory size. Pooled--Full TOST $p$-values are .0416, .0514, and .0008 at memories 200, 500, and 2,000. Equivalence is not established at the prespecified 500-memory anchor: its upper 90\% limit is $1.008$ pp. A separate test on the same units replaces the $N$ reference with the fixed anchor $a=-2$, held fixed across memory sizes: Full minus fixed is $-0.28$ pp (90\% CI $[-0.89,0.33]$) at memory 200 and $+0.12$ pp ($[-0.25,0.48]$) at 2,000, both within $\pm1$ pp after Holm correction over the two memories (Table~\ref{tab:v77-fixed-v6}).

\begin{table}[H]
\centering\small
\caption{\textbf{Deletion costs persist, but pooled-reference equivalence is unresolved at memory 500.} Eight paired units per memory. Differences in pp; 95\% intervals for deletion and 90\% intervals for equivalence against $\pm1$ pp. Full is shared with the correspondence sweep.}
\label{tab:v76-memory}
\begin{tabular}{@{}rcc@{}}\toprule
Memory & Full $-$ Delete $N$ [95\% CI] & Pooled $-$ Full [90\% CI]\\\midrule
200 & $9.129\ [8.059,10.198]$ & $0.344\ [-0.272,0.959]$\\
500 & $14.329\ [13.368,15.290]$ & $0.209\ [-0.591,1.008]$\\
2,000 & $11.741\ [10.991,12.491]$ & $0.121\ [-0.210,0.453]$\\\bottomrule
\end{tabular}
\end{table}

\begin{table}[H]\centering\small\setlength{\tabcolsep}{4pt}
\caption{Historical correspondence across memory budgets (eight paired units per memory). Each cell gives a difference in pp and a paired 95\% interval. Starred contrasts cross implementations.}
\label{tab:app-h-budgets}
\begin{tabular}{@{}rlccc@{}}\toprule
Memory & Comparison & $J$ & $A_{\rm task}$ & $C_{\rm task}$\\\midrule
200 & Full -- Iso H* & \shortstack{$+1.166$\\$[-0.919,+3.251]$} & \shortstack{$+1.397$\\$[-0.630,+3.425]$} & \shortstack{$+6.719$\\$[+5.111,+8.326]$} \\
200 & Iso -- Perm H* & \shortstack{$+1.450$\\$[+0.526,+2.374]$} & \shortstack{$+1.378$\\$[+0.437,+2.318]$} & \shortstack{$+6.619$\\$[+6.096,+7.142]$} \\
200 & Full -- Perm H & \shortstack{$+2.616$\\$[+1.075,+4.157]$} & \shortstack{$+2.775$\\$[+1.205,+4.345]$} & \shortstack{$+13.337$\\$[+11.781,+14.894]$} \\
\addlinespace
500 & Full -- Iso H* & \shortstack{$+3.326$\\$[+2.707,+3.945]$} & \shortstack{$+3.709$\\$[+3.075,+4.343]$} & \shortstack{$+4.929$\\$[+4.001,+5.857]$} \\
500 & Iso -- Perm H* & \shortstack{$+0.421$\\$[-0.410,+1.253]$} & \shortstack{$+0.027$\\$[-0.843,+0.898]$} & \shortstack{$+6.986$\\$[+6.166,+7.806]$} \\
500 & Full -- Perm H & \shortstack{$+3.747$\\$[+3.000,+4.495]$} & \shortstack{$+3.736$\\$[+2.794,+4.679]$} & \shortstack{$+11.915$\\$[+11.032,+12.798]$} \\
\addlinespace
2000 & Full -- Iso H* & \shortstack{$+3.119$\\$[+2.580,+3.658]$} & \shortstack{$+3.962$\\$[+3.151,+4.774]$} & \shortstack{$+3.303$\\$[+2.895,+3.710]$} \\
2000 & Iso -- Perm H* & \shortstack{$+0.580$\\$[-0.249,+1.409]$} & \shortstack{$+0.144$\\$[-0.907,+1.195]$} & \shortstack{$+5.336$\\$[+4.892,+5.781]$} \\
2000 & Full -- Perm H & \shortstack{$+3.699$\\$[+3.138,+4.260]$} & \shortstack{$+4.106$\\$[+3.419,+4.794]$} & \shortstack{$+8.639$\\$[+8.268,+9.009]$} \\
\addlinespace
\bottomrule\end{tabular}\end{table}

\begin{table}[H]\centering\small
\caption{Task-coordinate readouts of the pooling sweep, computed by the diagnostic pipeline (eight paired units per memory). Each cell gives a difference in pp and a paired 95\% interval, not an equivalence interval.}
\label{tab:app-n-budget-coordinates}
\begin{tabular}{@{}rlcc@{}}\toprule
Memory & Comparison & $A_{\rm task}$ & $C_{\rm task}$\\\midrule
200 & Full -- Delete N & \shortstack{$+10.259$\\$[+8.798,+11.719]$} & \shortstack{$-5.713$\\$[-6.579,-4.846]$} \\
200 & Pooled -- Full & \shortstack{$+0.316$\\$[-0.669,+1.302]$} & \shortstack{$+0.394$\\$[-0.227,+1.014]$} \\
\addlinespace
500 & Full -- Delete N & \shortstack{$+16.020$\\$[+14.957,+17.083]$} & \shortstack{$-4.895$\\$[-5.352,-4.438]$} \\
500 & Pooled -- Full & \shortstack{$+0.176$\\$[-0.860,+1.212]$} & \shortstack{$+0.260$\\$[-0.537,+1.057]$} \\
\addlinespace
2000 & Full -- Delete N & \shortstack{$+13.976$\\$[+13.105,+14.847]$} & \shortstack{$-4.019$\\$[-4.238,-3.799]$} \\
2000 & Pooled -- Full & \shortstack{$+0.260$\\$[-0.256,+0.776]$} & \shortstack{$-0.226$\\$[-0.573,+0.120]$} \\
\addlinespace
\bottomrule\end{tabular}\end{table}

\begin{table}[H]\centering\small
\caption{Older-task benefits coexist with latest-task costs across memory capacities (eight paired units per memory). Each cell gives a frozen-endpoint difference in pp and a descriptive paired 95\% interval, not an equivalence interval.}
\label{tab:app-n-budget-pop}
\begin{tabular}{@{}rlcc@{}}\toprule
Memory & Comparison & Older tasks & Latest task\\\midrule
200 & Full -- Delete N & \shortstack{$+10.664$\\$[+9.363,+11.965]$} & \shortstack{$-4.688$\\$[-8.199,-1.176]$} \\
200 & Pooled -- Full & \shortstack{$+0.424$\\$[-0.478,+1.325]$} & \shortstack{$-0.375$\\$[-1.690,+0.940]$} \\
\addlinespace
500 & Full -- Delete N & \shortstack{$+16.789$\\$[+15.548,+18.030]$} & \shortstack{$-7.812$\\$[-10.092,-5.533]$} \\
500 & Pooled -- Full & \shortstack{$+0.249$\\$[-0.952,+1.450]$} & \shortstack{$-0.150$\\$[-1.538,+1.238]$} \\
\addlinespace
2000 & Full -- Delete N & \shortstack{$+14.199$\\$[+13.255,+15.142]$} & \shortstack{$-10.375$\\$[-13.040,-7.710]$} \\
2000 & Pooled -- Full & \shortstack{$+0.046$\\$[-0.394,+0.486]$} & \shortstack{$+0.800$\\$[-0.906,+2.506]$} \\
\addlinespace
\bottomrule\end{tabular}\end{table}
\subsection{A refreshed later-class target: X-DER}\label{app:v8-xder}
X-DER rewrites the stored scores of classes learned after an example entered memory with the current model's responses \citep{bosch2022xder}. On eight further seeds (CIFAR-100, memory 500, ResNet-18, X-DER's published hyperparameters for this setting) we intercepted only the rewrite of the block of the task being learned: the published method (Full), the rewritten block replaced by its per-record mean after X-DER's rescaling (Flat), or no rewrite of that block (None). Later blocks are refreshed in all three arms. The prespecified primary test asked whether Full and Flat are equivalent within $\pm1$ pp; the Full$-$None and Flat$-$None contrasts were secondary.
\begin{table}[H]\centering\small
\caption{\textbf{A refreshed later-class target is not replaceable by its mean.} X-DER on CIFAR-100, eight paired seeds; final class-incremental accuracy and paired differences in pp. The primary Full$-$Flat test uses a 90\% interval (equivalence margin $\pm1$ pp); the secondary contrasts use 95\% intervals.}\label{tab:v8-xder}
\begin{tabular}{@{}lrrr@{}}\toprule
Contrast & Difference & CI & Positive/$n$\\\midrule
Full $-$ Flat (primary) & $+6.36$ & $[4.87,7.84]_{90\%}$ & 8/8\\
Full $-$ None & $+6.05$ & $[4.70,7.40]_{95\%}$ & 8/8\\
Flat $-$ None & $-0.31$ & $[-1.84,1.22]_{95\%}$ & 5/8\\\midrule
\multicolumn{4}{@{}l}{Arm means: Full 49.58, Flat 43.22, None 43.53 (seed SD 0.99, 1.90, 1.86).}\\\bottomrule
\end{tabular}
\end{table}
Equivalence fails: Full exceeds Flat by 6.36 pp, close to the 6.05 pp by which it exceeds None, and Flat does not differ detectably from not rewriting ($-0.31$ pp, 95\% CI $[-1.84,1.22]$). In the main study, the later-class scores are stored before those classes are learned, and selected fixed anchors replace them while meeting the $\pm1$ pp equivalence criterion on CIFAR-100. The scores X-DER writes come from a model that has learned those classes, and, as on $H$, their pattern carries information that their mean does not. The constant-reference result therefore applies to unrefreshed later-class targets. This is one cohort of eight seeds.

\section{Task-Level Offset Readouts}\label{app:cal}

\paragraph{Procedure.} For every replay model, logits on the 10{,}000 test images are recomputed from the saved final checkpoint; a model whose recomputed accuracy differs from its logged accuracy by more than 0.025 pp (at least three of 10{,}000 images) is excluded from every pair that uses it (three models in the ResNet-18 cohorts, one on ResNet-32). With $d(x)=\log\sum_{c\in T_{\rm last}}e^{z_c(x)}-\log\sum_{c\notin T_{\rm last}}e^{z_c(x)}$, the offset $b$ solves $\mathbb E_x[\sigma(d(x)+b)]=\pi$, where $\pi$ is the known share of the final task $T_{\rm last}$ among the class-stratified fitting images (no per-image labels enter the fit), by 80 steps of bisection. In the three withheld-image cohorts (ResNet-18, ResNet-32, and reduced-width ResNet-18), 20 training images per class (2{,}000, stratified and fixed across seeds and arms) are removed before training; $b$ is fitted on the first five per class (500) and, as a check, on all twenty, and the test set only scores. Elsewhere, unless stated otherwise, the test set is split into five class-stratified folds, and each fold is corrected with the offset fitted on the other four; the richer corrections of Appendix~\ref{app:cal-ladder} use two halves instead. In the 500-image check, five images per class fit $b$ and the remaining 9{,}500 are scored. In the distillation study, $b$ is fitted on the block of classes learned second, using 50 held-out training images per class that no student trains on, and applied to the saved test logits. Group-restricted accuracy (Section~\ref{sec:replay-evaluation}) and conditional cross-entropy restrict prediction to the true one of two groups ($T_{\rm last}$ or all earlier tasks; in distillation, the true half) and are invariant to $b$; they differ from the ten-task oracle $C_{\rm task}$. Differences are paired by seed, with paired $t$ intervals. P marks contrasts prespecified for that cohort; the others are descriptive.

\begin{table}[H]\centering\footnotesize\setlength{\tabcolsep}{3pt}
\caption{\textbf{Intervention costs before and after the task-level offset.} DER++ on CIFAR-100 (ResNet-18 unless stated) and TinyImageNet. Accuracy costs in pp with paired 95\% intervals; ``Block'' is the difference in group-restricted accuracy (pp) and ``Cond.\ CE'' in conditional cross-entropy (nats), both after the offset and unchanged by it. The fixed anchor uses $a=-2$. The first three cohorts fit the offset on 500 training images withheld from training; the others use five-fold test cross-fitting.}\label{tab:cal-interventions}
\resizebox{\linewidth}{!}{\begin{tabular}{@{}llrllrrc@{}}\toprule
Cohort & Contrast & $n$ & As trained & After offset & Block & Cond.\ CE & Spec.\\\midrule
Withheld-image cohort (Fig.~\ref{fig:offset}a) & Full $-$ Delete $N$ & 8 & $+14.89\ [13.81,15.97]$ & $+1.76\ [0.74,2.79]$ & $+1.99$ & $+0.04$ & P\\
 & Full $-$ Fixed & 8 & $+0.19\ [-0.47,0.85]$ & $+0.34\ [-0.13,0.80]$ & $+0.35$ & $-0.02$ & P\\
 & Full $-$ Perm $H$ & 8 & $+4.29\ [3.45,5.12]$ & $+3.98\ [3.47,4.49]$ & $+4.62$ & $-0.46$ & P\\
 & Perm $H-$ Delete $N$ & 8 & $+10.60\ [9.23,11.97]$ & $-2.22\ [-3.18,-1.25]$ & $-2.63$ & $+0.50$ & \\\addlinespace[2pt]
Withheld-image cohort, ResNet-32 & Full $-$ Delete $N$ & 8 & $+6.06\ [5.34,6.79]$ & $-1.55\ [-2.17,-0.93]$ & $-1.76$ & $-0.08$ & P\\
 & Full $-$ Fixed & 8 & $-0.49\ [-1.44,0.46]$ & $+0.39\ [-0.18,0.95]$ & $+0.01$ & $+0.04$ & P\\
 & Full $-$ Perm $H$ & 8 & $+3.45\ [2.61,4.28]$ & $+4.19\ [3.68,4.70]$ & $+4.41$ & $-0.52$ & P\\
 & Perm $H-$ Delete $N$ & 8 & $+2.62\ [1.67,3.56]$ & $-5.74\ [-6.26,-5.22]$ & $-6.17$ & $+0.43$ & \\\addlinespace[2pt]
Withheld-image cohort, reduced ResNet-18 & Full $-$ Delete $N$ & 8 & $+10.60\ [9.68,11.52]$ & $-0.59\ [-1.46,0.29]$ & $-0.37$ & $+0.05$ & P\\
 & Full $-$ Fixed & 8 & $-0.29\ [-0.90,0.31]$ & $-0.20\ [-0.80,0.41]$ & $-0.14$ & $-0.04$ & P\\
 & Full $-$ Perm $H$ & 8 & $+3.42\ [2.53,4.32]$ & $+3.19\ [2.21,4.17]$ & $+3.62$ & $-0.45$ & P\\
 & Perm $H-$ Delete $N$ & 8 & $+7.18\ [6.39,7.97]$ & $-3.78\ [-4.48,-3.07]$ & $-3.99$ & $+0.49$ & \\\addlinespace[2pt]
Paired cohort & Full $-$ Delete $N$ & 8 & $+14.01\ [13.27,14.76]$ & $+1.77\ [0.64,2.91]$ & $+1.76$ & $+0.04$ & \\
 & Full $-$ Fixed & 8 & $-0.49\ [-1.01,0.02]$ & $+0.31\ [-0.25,0.87]$ & $+0.22$ & $-0.03$ & P\\
 & Full $-$ Perm $H$ & 8 & $+3.92\ [3.19,4.65]$ & $+3.91\ [3.26,4.57]$ & $+4.53$ & $-0.46$ & P\\
 & Perm $H-$ Delete $N$ & 8 & $+10.09\ [9.15,11.03]$ & $-2.14\ [-3.11,-1.17]$ & $-2.77$ & $+0.50$ & \\\addlinespace[2pt]
Second cohort & Full $-$ Delete $N$ & 7 & $+14.99\ [13.40,16.59]$ & $+1.81\ [1.09,2.53]$ & $+2.03$ & $+0.04$ & \\
 & Full $-$ Fixed & 6 & $+0.31\ [-0.64,1.25]$ & $+0.32\ [-0.41,1.04]$ & $+0.28$ & $-0.02$ & \\
 & Full $-$ Perm $H$ & 8 & $+4.42\ [3.92,4.93]$ & $+4.22\ [3.61,4.83]$ & $+4.68$ & $-0.46$ & P\\
 & Perm $H-$ Delete $N$ & 7 & $+10.70\ [8.99,12.42]$ & $-2.31\ [-2.80,-1.83]$ & $-2.57$ & $+0.49$ & \\\addlinespace[2pt]
Fresh deletion cohort & Full $-$ Delete $N$ & 8 & $+14.54\ [14.17,14.90]$ & $+1.38\ [0.51,2.26]$ & $+1.71$ & $+0.06$ & P\\\addlinespace[2pt]
ResNet-32, cross-fitted & Full $-$ Delete $N$ & 19 & $+6.17\ [5.72,6.62]$ & $-1.97\ [-2.41,-1.53]$ & $-2.09$ & $+0.00$ & \\
 & Full $-$ Fixed & 19 & $-0.17\ [-0.74,0.40]$ & $+0.37\ [-0.11,0.85]$ & $+0.26$ & $+0.02$ & \\
 & Full $-$ Perm $H$ & 19 & $+3.74\ [3.33,4.14]$ & $+4.29\ [3.96,4.63]$ & $+4.71$ & $-0.55$ & \\\addlinespace[2pt]
TinyImageNet & Full $-$ Delete $N$ & 6 & $+8.28\ [6.36,10.20]$ & $-1.05\ [-1.88,-0.21]$ & $-0.84$ & $+0.00$ & \\\addlinespace[2pt]
DER & Full $-$ Perm $H$ & 5 & $+4.77\ [3.65,5.90]$ & $+4.45\ [3.82,5.08]$ & $+5.15$ & $-0.49$ & \\\addlinespace[2pt]
TinyImageNet, $H$ cohort & Full $-$ Perm $H$ & 3 & $+6.01\ [4.21,7.81]$ & $+5.92\ [4.68,7.15]$ & $+6.85$ & $-0.76$ & \\
 & Full $-$ $H$ block mean & 3 & $+3.85\ [2.76,4.93]$ & $+5.03\ [3.80,6.26]$ & $+4.86$ & $-0.68$ & \\\bottomrule
\end{tabular}}
\end{table}

In the withheld-image cohort, all three prespecified tests pass under Holm correction: the reassignment cost after the offset exceeds zero ($p<10^{-6}$), the offset reduces the deletion cost (by $13.13$ pp, $[11.44,14.81]$; $p<10^{-6}$), and Full and Fixed are equivalent after the offset (90\% CI $[-0.04,0.71]$, TOST $p=.006$). Fitting on all 2{,}000 withheld images changes every contrast by at most 0.05 pp, and the fitted offsets give the same contrasts as the best single offset for the test set to within 0.02 pp. The ResNet-32 withheld-image cohort repeats the holdout, recipe, arms, and tests with only the backbone changed, and all three prespecified tests pass under Holm correction: the reassignment cost after the offset exceeds zero ($+4.19$ pp, $p<10^{-6}$), the offset reduces the deletion cost (by $7.62$ pp, $[6.70,8.53]$; $p<10^{-6}$), and Full and Fixed are equivalent after the offset (90\% CI $[-0.07,0.84]$, TOST $p=.018$). Here the corrected deletion cost is negative ($-1.55$ pp): on this backbone the offset recovers more than the raw deletion cost, and Delete $N$ is the more accurate model after the offset. A third cohort repeats the protocol on a reduced-width ResNet-18 (the ResNet-18 block layout with fewer channels), and all three prespecified tests pass again: the reassignment cost after the offset exceeds zero ($+3.19$ pp, $[2.21,4.17]$; Holm $p=1.2\times10^{-4}$), the offset reduces the deletion cost (by $11.19$ pp, $[10.68,11.70]$; $p<10^{-6}$), and Full and Fixed are equivalent after the offset (90\% CI $[-0.68,0.29]$, TOST $p=.008$).

\paragraph{One offset per task in three model configurations.} The ResNet-32 and reduced-width (Mammoth's \texttt{reduced-resnet18}, stage widths 20/40/80/160 instead of 64/128/256/512) cohorts prespecified two secondary readouts: one offset per task, fitted by cross-entropy on the same 500 withheld images, and accuracy with prediction restricted to the true task. For the ResNet-18 cohort we computed both post hoc from its saved logits, which reproduce its single-offset accuracies exactly (Table~\ref{tab:cal-backbones}). The single-offset deletion cost changes sign across backbones, whereas with per-task offsets Delete $N$ is the more accurate model on all three, by 2.45 to 2.98 pp, and by 5.03 to 7.10 pp when prediction is restricted to the true task. The reassignment cost changes by at most 0.16 pp with the added offsets. What $N$ matching adds after one offset is therefore the allocation of predictions across tasks, which per-task offsets supply; within the true task it costs accuracy. Because the corrected deletion cost is itself small, the post-offset equivalence of Fixed and Full concerns the anchor's cost and does not, by itself, establish an advantage over Delete $N$: Fixed $-$ Delete $N$ after one offset is $+1.43$ ($[0.18,2.68]$), $-1.94$ ($[-2.40,-1.48]$), and $-0.39$ pp ($[-1.15,0.37]$) in the three configurations (secondary contrasts). The three cohorts are separate studies, so comparisons across backbones are descriptive.

\begin{table}[H]\centering\footnotesize\setlength{\tabcolsep}{2pt}
\caption{\textbf{Withheld-image cohorts under one offset and one offset per task.} DER++ on CIFAR-100, eight seeds per backbone; accuracy costs in pp with paired 95\% intervals. Both offsets are fitted on the same 500 withheld training images; the task oracle restricts prediction to the true task and is unchanged by any per-task offset. $^\dagger$Per-task offsets and task oracle computed post hoc; prespecified for the other two backbones.}\label{tab:cal-backbones}
\begin{tabular}{@{}llrrr@{}}\toprule
Contrast & Backbone & One offset & Offset per task & Task oracle\\\midrule
Full $-$ Perm $H$ & ResNet-18$^\dagger$ & $+3.98\ [3.47,4.49]$ & $+3.82\ [3.21,4.43]$ & $+12.38\ [11.15,13.60]$\\
 & ResNet-32 & $+4.19\ [3.68,4.70]$ & $+4.35\ [3.95,4.76]$ & $+13.41\ [12.95,13.87]$\\
 & Reduced ResNet-18 & $+3.19\ [2.21,4.17]$ & $+3.24\ [2.36,4.12]$ & $+11.14\ [10.50,11.79]$\\\addlinespace[2pt]
Full $-$ Delete $N$ & ResNet-18$^\dagger$ & $+1.76\ [0.74,2.79]$ & $-2.45\ [-3.39,-1.50]$ & $-5.03\ [-5.41,-4.64]$\\
 & ResNet-32 & $-1.55\ [-2.17,-0.93]$ & $-2.98\ [-3.52,-2.44]$ & $-7.10\ [-7.72,-6.47]$\\
 & Reduced ResNet-18 & $-0.59\ [-1.46,0.29]$ & $-2.54\ [-3.24,-1.84]$ & $-6.12\ [-6.89,-5.34]$\\\bottomrule
\end{tabular}
\end{table}

\paragraph{Shifted task proportions.} Reweighting the test images of this cohort so that the final task holds a share of 0.3 or 0.5 (no retraining), an offset refitted on the withheld images with the known share still substantially reduces the deletion cost (to $+0.83$ and $-0.05$ pp) and leaves a reassignment cost of $2.60$ and $1.56$ pp, which shrinks with the share of earlier tasks on which $H$ acts. An offset fitted for the original share and applied under an unknown shift lowers every model's accuracy, by 1.8--2.4 pp at 0.3 and 6.1--7.6 pp at 0.5, so the correction presumes known task proportions.

In the paired cohort, the fixed anchor meets the $\pm1$ pp criterion as trained (90\% CI within the band) and after the offset ($[-0.13,0.76]$), with Holm-adjusted TOST $p=.026$ and $.022$ over the two readouts. In the second cohort, two Fixed models failed the recomputation gate, leaving six pairs, fewer than the eight the test required; its intervals are reported descriptively. The prespecified correspondence tests (cost $>0$ as trained and after the offset) pass in both cohorts (Holm $p\le3\times10^{-6}$), and pooled over the sixteen seeds the cost after the offset is $+4.07$ pp ($[3.67,4.47]$). In the fresh deletion cohort, the prespecified tests that the offset reduces the deletion cost and repairs more than half of it pass (reduction $+13.15$ pp, $[12.36,13.95]$); fitting on 500 images gives a corrected cost of $+1.34$ pp ($[0.43,2.24]$). On ResNet-32, DER, and TinyImageNet the readouts are post hoc analyses of completed cohorts; for DER, the TinyImageNet deletion cohort, and the TinyImageNet $H$ cohort they use the models still available (five of eight, six of eight, and three of four seeds; the others were stored on a retired host).

\begin{table}[H]\centering\footnotesize\setlength{\tabcolsep}{3pt}
\caption{\textbf{The level--dispersion interaction depends on the prediction rule.} Fixed-anchor factorial cohorts (DER++, CIFAR-100, eight seeds each); pp with paired 95\% intervals. The raw interaction was prespecified in every cohort; the offset readout was prespecified in the second-anchor cohort (secondary) and in both fresh cohorts, where Both $<$ Neither after the offset was also a prespecified secondary test on ResNet-32 (one-sided $p=.0009$).}\label{tab:cal-factorial}
\resizebox{\linewidth}{!}{\begin{tabular}{@{}lllll@{}}\toprule
Cohort & $I_J$ as trained & $I_J$ after offset & Both$-$Neither as trained & Both$-$Neither after offset\\\midrule
ResNet-18, $a=-2$ & $+6.38\ [5.33,7.43]$ & $-0.42\ [-0.94,0.11]$ & $+14.23\ [13.39,15.07]$ & $+0.77\ [0.09,1.46]$\\
ResNet-18, $a=-4$ & $+7.01\ [5.67,8.35]$ & $+0.45\ [-0.36,1.25]$ & $+14.69\ [13.69,15.68]$ & $+0.58\ [-0.11,1.27]$\\
ResNet-18, $a=-2$, fresh (cohort of Fig.~\ref{fig:offset}b) & $+5.72\ [4.20,7.24]$ & $+1.23\ [0.29,2.17]$ & $+14.31\ [13.01,15.61]$ & $+1.81\ [0.86,2.76]$\\
ResNet-32, $a=-2$ & $+4.39\ [3.16,5.61]$ & $-0.71\ [-1.51,0.09]$ & $+6.86\ [6.29,7.43]$ & $-1.55\ [-2.27,-0.83]$\\
ResNet-32, $a=-2$, fresh (cohort of Fig.~\ref{fig:offset}b) & $+3.95\ [2.37,5.54]$ & $-1.09\ [-2.25,0.08]$ & $+7.35\ [6.26,8.43]$ & $-1.54\ [-2.28,-0.79]$\\\bottomrule
\end{tabular}}
\end{table}

The offset reduces the interaction by 4.5 to 6.8 pp in every cohort (all intervals exclude zero). A $\pm1$ pp equivalence of the corrected interaction to zero was prespecified in the fresh cohorts and holds in neither: the residual is positive on ResNet-18 ($+1.23$, 90\% CI $[0.48,1.98]$) and negative on ResNet-32 ($-1.09$, $[-2.02,-0.15]$).

\begin{table}[H]\centering\footnotesize\setlength{\tabcolsep}{3pt}
\caption{\textbf{Standard replay methods before and after the offset.} CIFAR-100, ResNet-18, memory 500, the principal recipe for every method (DER and DER++ with $\alpha=0.3$, DER++ with $\beta=0.5$, ER-ACE with its default cosine classifier); 32 paired seeds, no model excluded. pp with paired 95\% intervals. The six prespecified tests (the offset reduces each gap vs.\ ER, and removes more than half of it) all pass under Holm correction; the ER-ACE equivalence is secondary.}\label{tab:cal-methods}
\begin{tabular}{@{}lllrr@{}}\toprule
Contrast & As trained & After offset & Block & Cond.\ CE\\\midrule
DER++ $-$ ER & $+15.18\ [14.64,15.72]$ & $+5.70\ [5.25,6.14]$ & $+5.86$ & $-1.19$\\
DER $-$ ER & $+13.46\ [13.01,13.91]$ & $+4.85\ [4.45,5.24]$ & $+5.11$ & $-1.16$\\
ER-ACE $-$ ER & $+12.78\ [12.43,13.13]$ & $+0.27\ [-0.14,0.68]$ & $+0.96$ & $-0.83$\\
DER++ $-$ DER & $+1.72\ [1.08,2.36]$ & $+0.85\ [0.47,1.23]$ & $+0.75$ & $-0.03$\\
DER++ $-$ ER-ACE & $+2.40\ [1.84,2.95]$ & $+5.42\ [5.06,5.78]$ & $+4.89$ & $-0.36$\\\midrule
\multicolumn{5}{@{}l}{Accuracy as trained $\rightarrow$ after offset (\%): ER 21.8 $\rightarrow$ 36.0, ER-ACE 34.6 $\rightarrow$ 36.3,}\\
\multicolumn{5}{@{}l}{DER 35.2 $\rightarrow$ 40.9, DER++ 37.0 $\rightarrow$ 41.7.}\\\bottomrule
\end{tabular}
\end{table}

After the offset, ER-ACE $-$ ER meets the $\pm1$ pp criterion (TOST $p<.001$). Fitting on 500 images gives corrected gaps of $+5.65$, $+4.82$, and $+0.25$ pp for DER++, DER, and ER-ACE against ER, and five alternative 500-image splits change them by at most 0.07 pp.

\paragraph{Distillation study.} Teachers are ResNet-18 networks trained for 30 epochs on a random half $A$ of the CIFAR-100 classes and then on all classes; E0 is the teacher at the start of the second phase and E1 after its first epoch, when it reaches 34\% accuracy on the other half $B$ (and 0\% at E0). Each student is a ResNet-18 trained from scratch for 100 epochs on 50 labeled images per class with a label cross-entropy and, at weight $\lambda=1$ and temperature 1, the terms of $\mathrm{KL}(p_T\|p_S)=\mathrm{KL}(P_T(\cdot)\|P_S(\cdot))+\sum_{g\in\{A,B\}}P_T(g)\,\mathrm{KL}(q_T(\cdot\mid g)\|q_S(\cdot\mid g))$, where $P(g)$ is the probability mass of half $g$ and $q(\cdot\mid g)$ the distribution within it. Students of one teacher share initialization, data subset, and sampling order. The four comparisons of within-block matching with label-only training and with mass matching at E0 and E1 were prespecified on these eight fresh teachers and all pass under Holm correction ($p<10^{-6}$).

\begin{table}[H]\centering\footnotesize\setlength{\tabcolsep}{4pt}
\caption{\textbf{Distillation terms before and after the offset.} Accuracy gain over label-only training in pp with paired 95\% intervals (eight teachers); label-only students reach 37.7\%. The last column is the reduction in conditional cross-entropy relative to label-only training (nats).}\label{tab:cal-kd}
\begin{tabular}{@{}llllr@{}}\toprule
Stage & Student & As trained & After offset & Cond.\ CE\\\midrule
E0 & Mass & $-9.66\ [-10.34,-8.98]$ & $-0.13\ [-0.41,0.16]$ & $0.50$\\
 & Within & $+6.06\ [5.37,6.74]$ & $+6.36\ [5.64,7.08]$ & $0.79$\\
 & Both & $-0.93\ [-1.57,-0.30]$ & $+5.51\ [4.73,6.28]$ & $0.98$\\\addlinespace[2pt]
E1 & Mass & $+0.55\ [-0.02,1.11]$ & $+0.77\ [0.15,1.38]$ & $0.58$\\
 & Within & $+5.90\ [5.45,6.35]$ & $+5.23\ [4.70,5.76]$ & $0.99$\\
 & Both & $+7.20\ [6.77,7.63]$ & $+6.40\ [5.84,6.97]$ & $1.08$\\\bottomrule
\end{tabular}
\end{table}

After the offset, within-block matching exceeds mass matching by $+6.49$ pp ($[5.60,7.38]$) at E0 and $+4.47$ pp ($[3.85,5.08]$) at E1; adding mass to within-block matching changes accuracy by $-0.86$ pp ($[-1.60,-0.12]$) at E0 and $+1.17$ pp ($[0.75,1.59]$) at E1. Mass matching is within $\pm1$ pp of label-only training after the offset at E0 (TOST $p<.001$) but not at E1. Mass matching also lowers the conditional cross-entropy, so a loss term confined to one part of the target is not confined to that part of the prediction; the two terms share parameters.

\subsection{Content of the Within-Block Distillation Target}\label{app:kd-content}
A separate confirmation with eight fresh teachers (the same recipe, students, and stages) asks what in the within-half teacher distribution carries the gain of within-block matching (W). Three edited targets keep each record's teacher block mass $P_T(g)$ and its gold probability. Wperm cyclically shifts the non-gold conditional probabilities by one position over the non-gold classes of each half (all probabilities in a half without the gold class), so it preserves the set of values and the entropy and leaves no class on its own value. Wflat spreads the non-gold mass evenly, and Wtmpl replaces the non-gold proportions by the leave-one-record-out mean teacher conditional over training records of the same class. Neither edit is constrained to preserve entropy. WA and WB match the conditional only within the first-learned half $A$ or the second half $B$. The three comparisons of W with the edited targets at E1 were prespecified (one-sided, Holm over three) and all pass (Holm $p\le2\times10^{-4}$; Table~\ref{tab:kd-content}). Reassigned probabilities are worse than label-only training, and the class template recovers only part of the gain, so W uses record-specific information beyond this template; the template control does not establish that it carries none. At E0 nearly all of the gain comes from the conditionals of the old half $A$; at E1 both halves contribute, sub-additively.

\begin{table}[H]\centering\footnotesize\setlength{\tabcolsep}{4pt}
\caption{\textbf{The within-block gain needs class-specific, record-specific conditionals.} Accuracy differences after the offset in pp with paired 95\% intervals (eight teachers); label-only students reach 38.0\% after the offset. P marks the prespecified family (E1).}\label{tab:kd-content}
\begin{tabular}{@{}lllc@{}}\toprule
Contrast & E0 & E1 & Spec.\\\midrule
W $-$ label only & $+6.29\ [5.53,7.04]$ & $+5.21\ [4.19,6.22]$ & \\
W $-$ Wperm & $+7.50\ [6.97,8.03]$ & $+8.13\ [7.34,8.91]$ & P\\
W $-$ Wflat & $+6.84\ [6.28,7.39]$ & $+5.66\ [4.77,6.55]$ & P\\
W $-$ Wtmpl & $+4.11\ [3.60,4.62]$ & $+3.40\ [2.13,4.68]$ & P\\
Wperm $-$ label only & $-1.21\ [-1.97,-0.45]$ & $-2.92\ [-3.61,-2.24]$ & \\
Wflat $-$ label only & $-0.55\ [-1.12,0.02]$ & $-0.46\ [-1.21,0.30]$ & \\
Wtmpl $-$ label only & $+2.18\ [1.63,2.72]$ & $+1.80\ [0.95,2.65]$ & \\
WA $-$ label only & $+6.01\ [5.48,6.54]$ & $+3.06\ [2.33,3.80]$ & \\
WB $-$ label only & $-0.38\ [-1.04,0.28]$ & $+3.35\ [2.36,4.34]$ & \\
Interaction (W $-$ WA) $-$ (WB $-$ label only) & $+0.66\ [-0.18,1.49]$ & $-1.21\ [-2.21,-0.20]$ & \\\bottomrule
\end{tabular}
\end{table}

\subsection{A Text Check of the Distillation Decomposition}\label{app:kd-text}
We repeated the distillation study with DistilBERT \citep{sanh2019distilbert} on AG News \citep{zhang2015character}: four classes in two halves of two classes each. Teachers are fine-tuned for three epochs on half $A$ (E0) and then one epoch on all classes (E1), after which they reach 91 to 96\% on half $B$ across partitions (0\% at E0); students fine-tune the pretrained encoder with a new head for three epochs on 1{,}000 records per class, and the offset is fitted on 500 validation records per class. Four comparisons on the original partition were prespecified (within-block matching against label-only training and against mass matching at E0 and E1; one-sided, Holm over four; eight teachers), and two pass, both at E1. Two other partitions were then run as development cohorts with four teachers each. On every partition one offset removes most of the large raw cost of mass matching at E0, but the corrected within-block gain of the original partition does not appear under the other two, and under one of them within-block matching is below mass matching at E1 (Table~\ref{tab:kd-text}). Label-only students already reach about 90.6\%, and each half holds only two classes, so the non-gold reassignment of Appendix~\ref{app:kd-content} cannot be applied within the half that contains the gold class.

\begin{table}[H]\centering\footnotesize\setlength{\tabcolsep}{3pt}
\caption{\textbf{On text, offset repair transfers; within-block gains are small and depend on the partition.} DistilBERT on AG News; accuracy differences in pp, after the offset unless marked raw, with paired 95\% intervals. Halves: original \{World, Sports\} / \{Business, Sci/Tech\}; split 2 \{World, Business\} / \{Sports, Sci/Tech\}; split 3 \{World, Sci/Tech\} / \{Sports, Business\}. Numbers of teachers in parentheses; P marks the prespecified cohort, and the others are development cohorts.}\label{tab:kd-text}
\begin{tabular}{@{}llllll@{}}\toprule
 & \multicolumn{2}{l}{E0 mass $-$ label only} & E0 within $-$ & E1 within $-$ & E1 within $-$\\
Partition & raw & after & label only & label only & mass\\\midrule
Original (8, P) & $-9.37$ & $-0.13$ & $-0.12\ [-0.24,0.00]$ & $+0.28\ [0.17,0.38]$ & $+0.34\ [0.24,0.44]$\\
Split 2 (4) & $-7.90$ & $-0.43$ & $-0.09\ [-0.30,0.11]$ & $-0.03\ [-0.15,0.09]$ & $-0.20\ [-0.44,0.04]$\\
Split 3 (4) & $-6.59$ & $-0.40$ & $-0.16\ [-0.47,0.15]$ & $-0.12\ [-0.31,0.07]$ & $-0.45\ [-0.83,-0.08]$\\\bottomrule
\end{tabular}
\end{table}

\subsection{Richer Corrections}\label{app:cal-ladder}
A single task-level offset provides a minimal correction for a uniform shift in that task's logits. To test whether a cost that survives it would vanish under a more flexible correction, we refit, for the same trained models, one offset per task (nine free parameters, the first task fixed) and nine offsets together with a scale on the final task, each by minimizing cross-entropy on one half of the test images and scoring the other half, with the halves then swapped (the same split for every model). This richer-correction analysis is post hoc. The task oracle, accuracy with prediction restricted to the true task, is unchanged by any per-task offset. The added freedom changes the reassignment cost and the gaps of DER and DER++ over ER by less than 0.5 pp, whereas ER-ACE stays within 0.4 pp of ER after each correction in Table~\ref{tab:cal-ladder}. In the fixed-anchor factorial, per-task offsets remove the backbone-dependent sign of the single-offset interaction, and Both becomes less accurate than Neither on both backbones, with a within-task (task-oracle) cost of about 5 pp.

\begin{table}[H]\centering\footnotesize\setlength{\tabcolsep}{3pt}
\caption{\textbf{Costs and gaps under corrections of increasing freedom.} Accuracy differences in pp with root-paired 95\% intervals for the per-task offsets (numbers of seeds in parentheses). Columns: as trained; one final-task offset; one offset per task; per-task offsets and a final-task scale; accuracy with prediction restricted to the true task, which no per-task offset changes. The ResNet-18 reassignment row pools the two paired cohorts; the one-offset column here is fitted by cross-entropy on half-splits and so differs slightly from Tables~\ref{tab:cal-interventions} and~\ref{tab:cal-factorial}.}\label{tab:cal-ladder}
\begin{tabular}{@{}lrrrrr@{}}\toprule
Contrast & As trained & One offset & Per-task offsets & $+$ scale & Task oracle\\\midrule
Full $-$ Perm $H$ (ResNet-18, 16) & $+4.17$ & $+4.07$ & $+4.08\ [3.64,4.51]$ & $+4.05$ & $+11.93$\\
DER++ $-$ ER (32) & $+15.18$ & $+5.70$ & $+5.24\ [4.78,5.70]$ & $+5.62$ & $+3.41$\\
DER $-$ ER (32) & $+13.46$ & $+4.85$ & $+4.42\ [4.01,4.83]$ & $+4.83$ & $+2.85$\\
ER-ACE $-$ ER (32) & $+12.78$ & $+0.27$ & $+0.09\ [-0.33,0.50]$ & $+0.39$ & $-2.04$\\
\midrule
Interaction $I_J$, ResNet-18 (8) & $+5.72$ & $+1.21$ & $+0.28\ [-0.33,0.89]$ & $+0.36$ & $-1.78$\\
Interaction $I_J$, ResNet-32 (8) & $+3.95$ & $-1.08$ & $+0.44\ [-0.66,1.55]$ & $+0.56$ & $-2.17$\\
Both $-$ Neither, ResNet-18 (8) & $+14.31$ & $+1.79$ & $-1.86\ [-2.49,-1.23]$ & $-1.94$ & $-5.12$\\
Both $-$ Neither, ResNet-32 (8) & $+7.35$ & $-1.53$ & $-2.49\ [-3.22,-1.76]$ & $-2.44$ & $-5.26$\\
\bottomrule
\end{tabular}
\end{table}

\section{Prediction Errors of the Principal Interventions}\label{app:v9-moved-figures}
The two figures below decompose the principal interventions' prediction errors and show the original factorial with flat references (Generic on CIFAR-100, Record mean on TinyImageNet) under the raw prediction rule.

\begin{figure}[H]
\centering
\includegraphics[width=\linewidth]{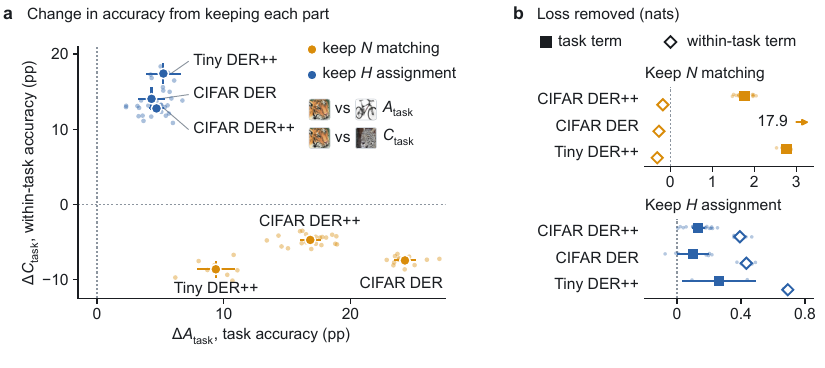}
\caption{\textbf{$N$ matching primarily supports task selection, whereas $H$ assignment primarily supports within-task discrimination.} (a) Change in task accuracy $A_{\rm task}$ and within-task accuracy $C_{\rm task}$: Full minus Delete $N$ (amber) and Full minus reassigned $H$ (blue), with paired seeds and marginal 95\% intervals; the image pairs illustrate the decision each accuracy scores. (b) The exact split of Equation~\ref{eq:v6-task-within-ce}: loss removed by keeping $N$ matching (top) and the $H$ assignment (bottom), on different scales; the DER task-term reduction, 17.9 nats, lies beyond the axis. Seeds (CIFAR-100/DER/TinyImageNet): 20/12/8 for deletion, 20/8/4 for reassignment. Keeping $N$ matching reduces the task term by 1.77 nats while the within-task term rises by 0.17; keeping the $H$ assignment reduces mainly the within-task term, which accounts for 61\%--85\% of the loss reduction across the nine $H$ comparisons (Table~\ref{tab:v77-loss-split-v6}).}
\label{fig:profiles}\label{fig:v77-diagnostics}\label{fig:v75-profiles}
\end{figure}

\begin{figure}[H]
\centering
\includegraphics[width=\linewidth]{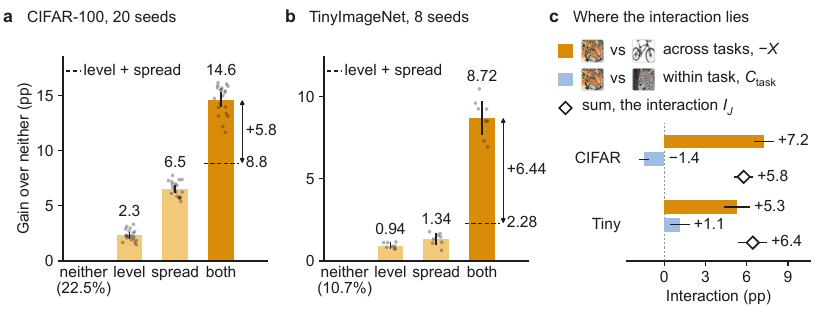}
\caption{\textbf{Under the raw prediction rule, both retained components outperform either alone.} (a,b) Accuracy gain of each policy over Delete $N$ (``neither''; ``spread'' is the dispersion term), with seeds and 95\% paired intervals; the dashed line is the sum of the two single-term gains and the arrow the interaction $I_J$. ``Both'' uses Generic on CIFAR-100 (20 seeds) and Record mean on TinyImageNet (8 seeds). (c) Since $J=C_{\rm task}-X$, $I_J$ is the sum of a cross-task part ($-X$) and a within-task part ($C_{\rm task}$), shown with 95\% intervals; the image pairs illustrate the two error types. On eight CIFAR-100 seeds, Both exceeds doubled Level only by 12.5 pp and doubled Dispersion only by 7.3 (Table~\ref{tab:v77-double-v6}).}
\label{fig:factorial}\label{fig:v75-factorial}
\end{figure}

\end{document}

%% file: authors.tex
\author{\parbox{0.96\textwidth}{\normalfont\centering
\normalsize BoRen Deng\textsuperscript{1}\quad Xiangyue Ma\textsuperscript{2}\quad Chenglong Li\textsuperscript{3}\quad Xiaoting Du\textsuperscript{1,*}\endgraf\vspace{4pt}
\small \textsuperscript{1}Beijing University of Posts and Telecommunications\endgraf \textsuperscript{2}Beihang University\quad \textsuperscript{3}Beijing University of Technology\endgraf
\vspace{3pt}\scriptsize BoRen Deng: \href{mailto:borendeng427@163.com}{borendeng427@163.com}\quad Xiangyue Ma: \href{mailto:xiangy_ma@buaa.edu.cn}{xiangy\_ma@buaa.edu.cn}\endgraf Chenglong Li: \href{mailto:lichenglong@bjut.edu.cn}{lichenglong@bjut.edu.cn}\quad Xiaoting Du: \href{mailto:duxiaoting@bupt.edu.cn}{duxiaoting@bupt.edu.cn}\endgraf
\vspace{3pt}\footnotesize *Corresponding author: Xiaoting Du}}